\documentclass[10pt,journal,compsoc]{IEEEtran}
\ifCLASSOPTIONcompsoc
  \usepackage[nocompress]{cite}
\else
  \usepackage{cite}
\fi
\ifCLASSINFOpdf
\else
\fi
\usepackage{graphicx}
\usepackage{romannum}
\usepackage{array,bbm}
\usepackage{dsfont}
\usepackage{subcaption}
\usepackage{layouts}
\usepackage{xcolor}
\usepackage{algorithm}
\usepackage[noend]{algpseudocode}
\usepackage{transparent}
\usepackage{setspace}
\graphicspath{{Gallery/}}

\usepackage{amsmath,amssymb}
\newtheorem{theorem}{Theorem}

\newtheorem{proposition}[theorem]{Proposition}

\newtheorem{definition}{Definition}
\begin{document}
%
\title{Deep Sparse Coding Using Optimized Linear Expansion of Thresholds}
%
%
%
%

\author{Debabrata Mahapatra, Subhadip Mukherjee, and Chandra Sekhar Seelamantula,~\IEEEmembership{Senior Member, IEEE}
\thanks{The authors are with the Indian Institute of Science, Department of Electrical Engineering, Bangalore - 560 012, India. Telephone: +91 80 2293 2695; Fax: +91 80 2360 0444; Email: chandra.sekhar@ieee.org, subhadip@ee.iisc.ernet.in, mahapatradebabrata91@gmail.com. This manuscript was submitted to IEEE Transactions on Pattern Analysis and Machine Intelligence (Manuscript ID: TPAMI-2016-11-0861; Submission date: November 11, 2016).}}

\markboth{}%
{Shell \MakeLowercase{\textit{et al.}}: Bare Demo of IEEEtran.cls for Computer Society Journals}
%



\IEEEtitleabstractindextext{%
\begin{abstract}
We address the problem of reconstructing sparse signals from noisy and compressive measurements using a feed-forward deep neural network (DNN) with an architecture motivated by the \textit{iterative shrinkage-thresholding algorithm} (ISTA). We maintain the weights and biases of the network links as prescribed by ISTA and model the nonlinear activation function using a \textit{linear expansion of thresholds} (LET), which has been very successful in image denoising and deconvolution. The optimal set of coefficients of the parametrized activation is learnt over a training dataset containing measurement-sparse signal pairs, corresponding to a fixed sensing matrix. For training, we develop an efficient second-order algorithm, which requires only matrix-vector product computations in every training epoch (Hessian-free optimization) and offers superior convergence performance than gradient-descent optimization. Subsequently, we derive an improved network architecture inspired by FISTA, a faster version of ISTA, to achieve similar signal estimation performance with about 50\% of the number of layers. The resulting architecture turns out to be a \textit{deep residual network}, which has recently been shown to exhibit superior performance in several visual recognition tasks. Numerical experiments demonstrate that the proposed DNN architectures lead to $3$-$4$ dB improvement in the reconstruction signal-to-noise ratio (SNR), compared with the state-of-the-art  sparse coding algorithms.
\end{abstract}

\begin{IEEEkeywords}
Compressive sensing, sparse recovery, iterative shrinkage, ISTA, FISTA, Hessian-free optimization, deep learning, back-propagation, linear expansion of thresholds (LET).
\end{IEEEkeywords}}

\maketitle

\IEEEdisplaynontitleabstractindextext

%
\IEEEpeerreviewmaketitle

\IEEEraisesectionheading{\section{Introduction}
\label{sec:introduction}}

%
%
%
%
\IEEEPARstart{E}{stimation} of sparse signals from inaccurate linear measurements, formally known as \textit{compressive sensing} (CS) \cite{candesintro2008,baraniuk1,candes_tao, donoho_it}, has gained enormous importance in signal processing over the past decade. Most signals occurring in real-world applications admit a sparse representation in an appropriate  basis. A significant amount of research has gone into designing suitable measurement matrices, and in developing efficient algorithms to reconstruct signals from noisy and incomplete set of measurements, based on the premise of sparsity. The problem is also equivalent to that of \textit{best basis selection} or \textit{sparse coding}, wherein one intends to find the sparse representation of a signal in an overcomplete set of basis vectors. Compressive sensing techniques have been successfully employed in a variety of signal processing applications such as deconvolution \cite{beck2009fast}, denoising \cite{elad2006image}, super-resolution \cite{yang2010image}, inpainting, etc.. The applications of CS have transcended the domain of classical signal processing and gone much beyond. For example, applications such as medical imaging \cite{lustig}, computational biology \cite{sheikh}, radar imaging \cite{comp_radar}, pattern classification\cite{wright2009robust}, feature extraction \cite{hyvarinen1998image}, etc. have immensely benefitted from the developments in CS.\\ 
\indent The objective in CS is to recover a sparse signal $\bold x\in \mathbb{R}^n$ from a lower-dimensional measurement vector $\bold y\in \mathbb{R}^m$, $m<n$, given by $\bold y = \bold A \bold x+\boldsymbol \xi$, where $\bold A$ is the sensing matrix and $\boldsymbol \xi$ is the measurement noise. Without assuming a prior on $\bold x$, the task of signal reconstruction is ill-posed. The goal of CS algorithms is to seek the sparsest $\bold x$ that is consistent with the measurement $\bold y$. However, searching over the space of all possible patterns of sparsity is a combinatorial problem and prohibitively expensive. In a seminal work \cite{candes_romberg_tao}, Cand\`es et al. showed that  it is possible to obtain a stable estimate of a sparse signal $\bold x$ with $s$ nonzero entries, by solving a tractable convex program, provided that $\left\|\boldsymbol \xi\right\|_2\leq \epsilon$ and $\bold A$ satisfies $\delta_{3s}+3\delta_{4s}<2$, where $\delta_s$ denotes the \textit{restricted isometry constant} (RIC) of order $s$. The convex programming formulation in \cite{candes_romberg_tao} is also known as \textit{basis pursuit denoising}.\\
\indent Recently, research on supervised learning of a deep neural network (DNN) for approximating fairly complex nonlinear functions has gained significant momentum. Because of the availability of abundant training data and enhanced computational capability, DNN-based algorithms outperform many classical techniques in applications related to natural language processing, speech processing, computer vision, etc., to name a few. For example, convolutional neural network-based models \cite{krizhevsky2012imagenet} have resulted in significant improvement in accuracy for the tasks of object recognition \cite{he2015deep}, detection \cite{sermanet2013overfeat}, and image segmentation \cite{DBLP:journals/corr/ChenPKMY14}. Deep learning architectures such as the restricted Boltzmann machine \cite{hinton2010practical}, auto-encoders \cite{rifai2012disentangling}, and recurrent neural networks \cite{sutskever2013training} are used extensively in automatic speech generation \cite{ling2015deep} and recognition \cite{graves2013speech} systems. Several DNN-based methods have also been proposed for unsupervised feature extraction \cite{rifai2012generative, vincent2010stacked}. Specifically, the spike-and-slab model \cite{goodfellow2012large} has been shown to work efficiently for sparse feature extraction from large-scale data.\\
\indent Sparse coding can be interpreted as a function approximation problem, where the goal is to undo the effect of the sensing matrix $\bold A$ to estimate $\bold x$ from $\bold y$, subject to the constraint of sparsity. Consequently, it is natural to ask how much sparse coding can benefit from a trained DNN, especially in applications where sufficient number of training examples are available. Gregor and Le Cun showed, for the first time, that a custom-designed discriminatively trained DNN architecture can solve the sparse coding problem efficiently \cite{lecun1}. Their framework is referred to as the {\it learned iterative shrinkage and thresholding algorithm (ISTA)}. Further efforts have been made to make a one-to-one correspondence between iterative inferencing algorithms and DNNs \cite{unfolding1}, and to efficiently represent the nonlinearity that is at the heart of sparse coding \cite{7442798}. Continuing along the same line of investigation, we show that a custom-designed, discriminatively trained DNN architecture can solve the sparse coding problem efficiently. We next give an overview of deep-learning-based algorithms for sparse coding available in the literature, before highlighting our contribution.
\subsection{Prior Art}
\indent The use of a trained NN for sparse coding is primarily based on the observation that the updates in an iterative algorithm can be interpreted as the layers of a deep neural network. This unfolding process has been applied in the context of ISTA \cite{ista_1}. The possibility of using a trained feed-forward neural network for efficient sparse coding was first demonstrated by Gregor and LeCun \cite{lecun1}. Their model, termed as \textit{learned ISTA}, implements a truncated version of ISTA with trained weights and biases instead of precomputed ones. The parameters of the network are trained over a dataset of measurement-signal pairs by minimizing the estimation error. However, the number of parameters to be trained in the learned ISTA model scales as $n^2$, where $n$ is the dimension of the sparse signal to be estimated. Hershey et al. \cite{unfolding1} proposed the idea of \textit{deep unfolding}, wherein an iterative inference strategy, such as ISTA, inspires the architecture of a DNN. The model parameters are untied across the layers and trained discriminatively using gradient-descent (GD). The idea of unfolding an iterative inference algorithm with shared parameters over the layers was developed by Domke \cite{domke1, domke2}, in the context of tree-reweighted belief propagation and mean-field inference. Sprechmann et al. \cite{sprechmann_sapiro} proposed a learnable architecture of fixed complexity, by unfolding the iterative proximal descent algorithms for structured sparse and low-rank models. The framework is termed as \textit{process-centric parsimonious modeling} and is capable of achieving similar or better performance than model-centric iterative approaches, while offering significant reduction in complexity. Similar ideas based on unfolding were exploited by Wang et al. \cite{wang_l0_encoder} to design a deep $\ell_0$ encoder, with applications in image classification and clustering. Recently, Xin et al. \cite{xin_max_sparsity} critically analyzed the merits of designing trainable models over conventional optimization-based sparse coding algorithms. They considered different architectures that improve the effective RIC, thereby successfully alleviating the issue of \textit{disruptive correlation} in the dictionary, which might cause standard sparse coding techniques to fail. Kamilov and Mansour \cite{7442798} proposed a deep architecture for sparse coding motivated by ISTA, wherein the nonlinear activation function is modeled using cubic B-splines. Their approach has at least two advantages: (i) the reconstruction accuracy is better than that of ISTA; and (ii) the number of parameters to be learnt does not scale with the signal dimension $n$. However, we shall show, among other things, that the choice of the basis functions could be optimized, which leads to a more parsimonious representation of the thresholding function. 

\subsection{This Paper}
\indent Akin to Kamilov and Mansour, we construct a DNN architecture for solving the sparse coding problem by unfolding the ISTA iterations (Section~\ref{IS_NN_connection_sec}). However, we parameterize the nonlinear activation function using a \textit{linear expansion of thresholds} (LET)\cite{pan2013iterative} instead of cubic B-splines (Section~\ref{sec_LET_prop}). The resulting network is referred to as the \textit{LETnet}. The weights and biases in each layer of \textit{LETnet} are precomputed following the ISTA prescription and kept fixed, while the parametrized activation function is fine-tuned to fit the training data (Section~\ref{LETnet_architecture_sec}). We show that such a model has lesser number of parameters to train, without compromising the learning ability of the network. Unlike \cite{7442798}, the coefficients of the LET-based activations are untied across the layers of the network to enhance the expressive power of the model, without significantly affecting the training overhead. Also, a small number of coefficients, typically five per layer, suffice, which results in considerably less number of parameters to learn overall in comparison with \cite{7442798}. The \textit{LETnet} architecture is trained discriminatively over a dataset to optimally tune the parameters for a given sensing matrix. We demonstrate that the LET-based activation function induces a variety of sparsity promoting regularizers and optimum trade-off between noise rejection and signal sparsity can be achieved in every layer by learning the LET coefficients appropriately. The flexibility to design and train a parametric family of regularizers to encourage sparsity, while effectively suppressing noise using a parsimonious parameterization is one of the major advantages of our approach. We also derive an efficient Hessian-free optimization (HFO) technique \cite{martens2010deep} to train the network (Section \ref{derivation_letnetvar}). Numerical experiments on synthesized signals indicate that the trained \textit{LETnet} is capable of producing an improvement in signal-to-noise ratio (SNR) of approximately 3 to 4 dB over the competing techniques (Section~\ref{num_valid_letnet_sec}).\\
\indent Subsequently, we also show that it is possible to further reduce the number of layers in the \textit{LETnet}, without compromising the recovery performance. The motivation is derived from the \textit{fast iterative shrinkage threshold algorithm} (FISTA) \cite{beck2009fast}, which has a faster convergence rate than ISTA without increasing the computational load per iteration, thereby requiring considerably less number of iterations overall. We show that the resulting fast \textit{LETnet} architecture, dubbed as \textit{fLETnet} (Section~\ref{fLETnet_architecture_sec_intro}), bears close resemblance to the recently proposed \textit{deep residual learning} architecture \cite{he2015deep}, wherein each layer draws direct connections from two preceding layers instead of one, so as to improve the convergence performance. It has been shown in \cite{he2015deep} that the scheme enjoys the merits of a deep architecture, while successfully alleviating the problems of vanishing/exploding gradients. This advantage comes without the need to learn additional parameters. We carry out experimental validation of \textit{LETnet} and \textit{fLETnet} for sparse signal recovery and demonstrate their superiority over the the learning-based approach proposed in \cite{7442798} as well as the conventional sparse coding algorithms that are not set up within a learning paradigm. 

\section{Iterative Shrinkage Algorithms: A Neural Network Perspective}
\label{IS_NN_connection_sec}
\indent Compressive sensing deals with the recovery of a sparse signal $\bold x \in \mathbb{R}^n$ from a measurement of the form
\begin{equation}
\bold y= \bold A \bold x +  \boldsymbol \xi,
\label{CS_measurement}
\end{equation}
where $\bold y \in \mathbb{R}^m$, $m<n$, $\bold A \in \mathbb{R}^{m \times n}$ is the sensing matrix and $ \boldsymbol \xi$ denotes the measurement noise. The signal $\bold x$ could be sparse itself or it could admit a sparse representation in an appropriate basis $\bold B \in \mathbb{R}^{n \times n}$, meaning that $\bold x=\bold B \boldsymbol \alpha$, where $\boldsymbol \alpha \in \mathbb{R}^{n}$ is sparse. For example, natural images can be represented sparsely if $\bold B$ is taken as the orthogonal discrete cosine transform or wavelet basis. Without loss of generality, we consider sparsity of $\bold x$ in the canonical basis (columns of an identity matrix). If $\bold x$ admits a sparse representation in a basis $\bold B$ different from the identity, $\bold A$ and $\bold x$ in \eqref{CS_measurement} should be replaced with $\bar{\bold A}=\bold A \bold B$ and $\boldsymbol \alpha$, respectively.\\
\indent In order to recover $\bold x$ from $\bold y$, one is required to solve the combinatorially hard optimization 
\begin{equation}
\hat{\bold x}=\arg\underset{\bold x}{\min}\left\|\bold x\right\|_0 \text{\,\,subject to\,\,}\left\|  \bold y-\bold A \bold x \right\|_2 \leq \epsilon,
\label{CS_main_eq}
\end{equation}
where the $\ell_0$-norm of $\bold x$, denoted by $\left\|\bold x\right\|_0$, counts the number of nonzero entries in $\bold x$. Solving \eqref{CS_main_eq} using an exhaustive search over all possible sparsity patterns becomes computationally intractable as it grows exponentially with $n$. There are two distinct algorithmic paradigms to circumvent this problem: (i) greedy algorithms; and (ii) relaxation-based approaches.\\
\indent In greedy algorithms, one first estimates the support of $\bold x$ using a greedy iterative approach, and subsequently computes a least-squares (LS) estimate over the estimated support to obtain the amplitudes. Algorithms such as orthogonal matching pursuit (OMP) \cite{tropp_omp}, compressive sampling matching pursuit (CoSaMP) \cite{tropp_cosamp}, subspace pursuit \cite{dai_subspace_pursuit}, etc. and their several variants, fall under this category.\\
\indent In relaxation-based algorithms, one minimizes a regularized quadratic penalty of the form
\begin{equation}
\underset{\bold x}{\min}\text{\,}\frac{1}{2}\left\|  \bold y-\bold A \bold x \right\|_2^2+\lambda \mathcal{G}\left(\bold x\right),
\label{CS_main_eq_lasso}
\end{equation}
where the regularizer $\mathcal{G}\left(\bold x\right)$ allows one to incorporate priors about $\bold x$. In this case, $\mathcal{G}$ is chosen appropriately to promote sparsity in the estimate, which stabilizes the otherwise ill-posed inverse problem. In the special case where $\mathcal{G}\left(\bold x\right)=\left\| \bold x\right\|_1$, the resulting optimization is referred to as LASSO regression \cite{tibshirani,tibshirani_new}. The parameter $\lambda>0$ trades-off between the signal prior and data fidelity.
\subsection{Proximal Gradient Methods}
\label{sec_prox_grad_explain}
\indent To make the exposition self-contained, we briefly recall the connection between the proximal gradient methods and the feed-forward neural networks (NN), originally established by Gregor and LeCun \cite{lecun1}. Consider an iterative algorithm for solving \eqref{CS_main_eq_lasso}, which generates an estimate $\bold x^{t}$ in the $t^{\text{th}}$ iteration. Denoting $f\left(\bold x\right)=\frac{1}{2}\left\|  \bold y-\bold A \bold x \right\|_2^2$, the affine approximation $f_t\left(\bold x\right)$ of $f\left(\bold x\right)$ at $\bold x^t$ is given by
\begin{equation}
f_t\left(\bold x\right)=f\left(\bold x^t\right)+\left( \bold x - \bold x^t\right)^\top\nabla f\left(\bold x^t\right).
\label{first_order_approx}
\end{equation}
The GD algorithm for minimizing $f\left(\bold x\right)$, with a fixed step-size $\eta$ at every $t$, takes the form
\begin{equation*}
\bold x^{t+1}=\bold x^t-\eta \nabla f\left(\bold x^t\right).  
\end{equation*}
The GD update corresponds to the minimization of a quadratically-regularized affine approximation:
\begin{equation*}
\bold x^{t+1}=\arg  \underset{\bold x}{\min}\text{\,\,}f_t\left(\bold x\right)+\frac{1}{2\eta}\left\| \bold x - \bold x^t \right\|_2^2.
\end{equation*}
Adopting a similar approximation for solving \eqref{CS_main_eq_lasso} leads to an iterative update of the form 
\begin{equation*}
\bold x^{t+1}=\arg  \underset{\bold x}{\min}\text{\,\,}f_t\left(\bold x\right)+\frac{1}{2\eta}\left\| \bold x - \bold x^t \right\|_2^2+\lambda \mathcal{G}\left(\bold x\right),
\end{equation*}
which, after rearranging the terms, can be written as 
\begin{equation}
\bold x^{t+1}=\arg  \underset{\bold x}{\min}\text{\,\,}\frac{1}{2\eta}\left\| \bold x - \left(\bold x^t-\eta \nabla f\left(\bold x^t\right)\right) \right\|_2^2+\lambda \mathcal{G}\left(\bold x\right).
\label{eq_prox_op}
\end{equation}
If $\mathcal{G}\left(\bold x\right)$ is separable, that is, if $\mathcal{G}\left(\bold x\right)=\sum_{i=1}^{n}\mathcal{G}_i\left(x_i\right)$, where $x_i$ denotes the $i^{\text{th}}$ entry of $\bold x$, then solving \eqref{eq_prox_op} boils down to solving $n$ independent 1-D optimization problems: 
\begin{equation*}
x^{t+1}_i=\arg  \underset{x_i}{\min}\text{\,}\frac{1}{2\eta}\left|  x_i- u^t_i\right|^2+\lambda \mathcal{G}_i\left(x_i\right), i = 1, 2, \cdots, n,
\label{eq_prox_op1}
\end{equation*}
where $\bold u^t=\bold x^t-\eta \nabla f\left(\bold x^t\right)$. Defining the proximal operator
\begin{equation}
P_{\nu}^{g}\left(u\right)\stackrel{\Delta}{=}\arg  \underset{x}{\min}\text{\,}\frac{1}{2}\left |  x-u \right |^2+\nu g(x), 
\label{prox_op_def}
\end{equation}
for $g(\cdot)$ corresponding to the regularization parameter $\nu$, and assuming $R_i \equiv g$ for every $i$, one can write 
\begin{equation}
\bold x^{t+1}=P_{\nu}^{g}\left(\bold x^t-\eta \nabla f\left(\bold x^t\right)\right),
\label{eq: Proximal_op_definition}
\end{equation}    
where $\nu=\lambda \eta$, and the proximal operator is applied coordinate-wise on its argument. In the special case, albeit an important one, where $g(x)=|x|$, the proximal operator turns out to be the soft-thresholding (ST) function $T_{\nu}\left(v\right)=\text{sgn}(v)\max\left \{ |v|-\nu,0 \right\}$, and the resulting update rule becomes the popular ISTA. Using the expression $\nabla f\left(\bold x\right)=\bold A^\top\left(\bold A \bold x- \bold y\right)$, \eqref{eq: Proximal_op_definition} is rewritten as 
\begin{equation}
\bold x^{t+1}=P_{\nu}^{g}\left(  \bold W \bold x^t +\bold b\right),
\label{eq_prox_op3}
\end{equation}
where $\bold W=\bold I - \eta \bold A^\top \bold A$ and $\bold b=\eta  \bold A^\top \bold y$. Gregor and LeCun \cite{lecun1} interpreted \eqref{eq_prox_op3} as the feed-forward computation through a NN with weight matrix $\bold W$ and bias $\bold b$ shared across every layer. The input to the NN is the initial estimate $\bold x^0$. The proximal operator $P_{\nu}^{g}\left(  \cdot \right)$ plays the role of nonlinear activation function of the neurons in layer $t$. An estimate obtained after $L$ iterations of \eqref{eq_prox_op3} is the same as the output of the NN containing $L$ layers. Thus, there is a direct correspondence between ISTA and a feed-forward NN.
\begin{figure*}[t!]
	\centering
	\captionsetup[subfigure]{labelformat=empty}
	\begin{subfigure}{.33\textwidth}
		\centering
		\includegraphics[]{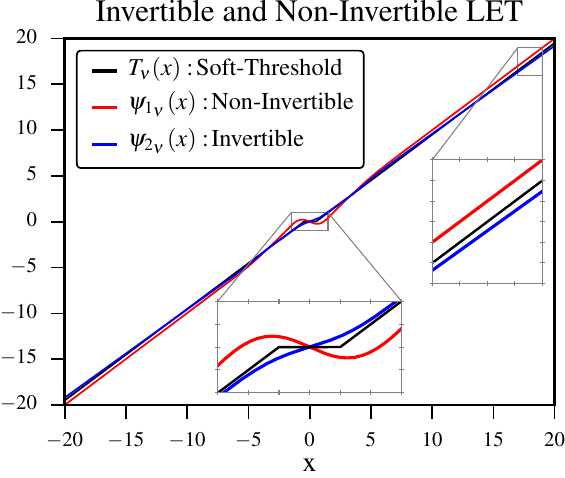}
		\caption{\; \; \; (a)}
		\label{sub_fig: LET}
	\end{subfigure}%
	\hspace{-2.5mm}
	\begin{subfigure}{.33\textwidth}
		\centering
		\includegraphics[]{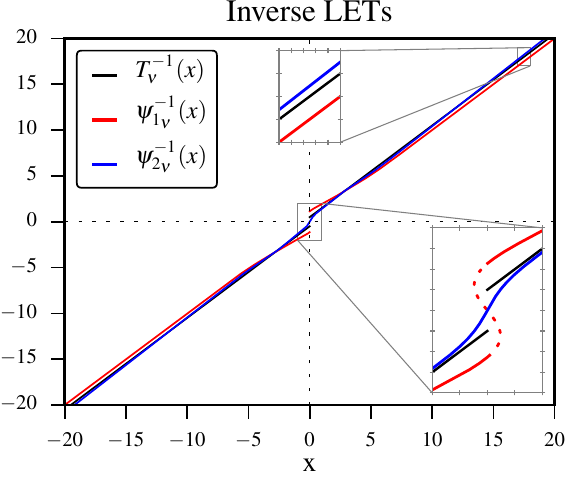}
		\caption{\; \; \; (b)}
		\label{sub_fig: inv}
	\end{subfigure}%
	\begin{subfigure}{.33\textwidth}
		\centering
		\includegraphics{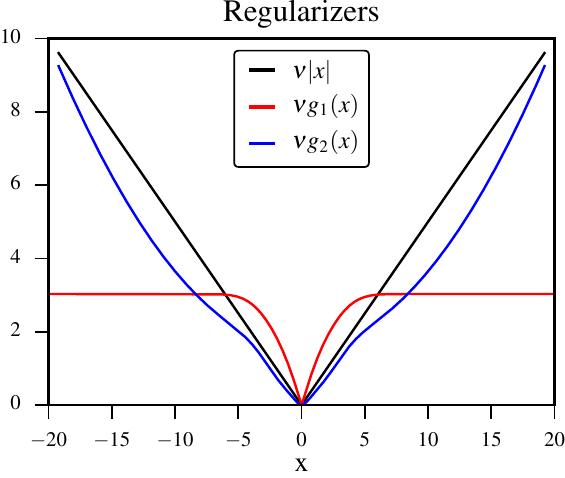}
		\caption{\; \;  (c)}
		\label{sub_fig: reg}
	\end{subfigure}%
	
	\caption{\small(Color online) Illustration of LET-based activation functions and the ST operator, together with the corresponding regularizers;\; \textbf{(a)} Both LET functions (in red and blue) are constructed using $K=4$ LET bases. The coefficients for the invertible (blue) and the non-invertible (red) LETs are \{$0.96,\, -0.87,\, 1.9,\, -1.58$\}  and  \{$1,\, -0.9,\,  1.3,\, -2$\}, respectively. The parameter $\tau$ is such that $3\tau = \nu$ with $\nu=0.5$; \textbf{(b)} inverse of the LET-based activation functions according to definition (\ref{eq: inverse_let_def});\; \textbf{(c)} the induced regularizers.
		\hfill}
	\label{fig: LET_inv_Reg}
\end{figure*}
\section{Parametric Activation: Linear Expansion of Thresholds (LET)}
\label{sec_LET_prop}
For convenience, we denote the NN analogue of the generic proximal operator $P_{\nu}^{g}$, namely the activation function, as $\psi$. Our approach concerns the construction of $\psi$ by taking a linear combination of $K$ elementary thresholding functions:
\begin{equation}
\psi(u) = \sum_{k = 1}^{K} c_k \phi_k(u), u\in \mathbb{R}.
\label{eq_let_activation_def}
\end{equation}
This design offers more flexibility than the ST operator $T_{\nu}$, because, by choosing a moderately small number (typically less than $10$) of appropriate elementary thresholding functions $\phi_k$, one could construct a rich variety of activations. We advocate the use  of elementary functions based on the \textit{derivatives of a Gaussian} (DoG) \cite{blu_surelet, pan2013iterative}, specified as
\begin{align}
\phi_k(u) =  u \, \exp\left(- \frac{(k-1)u^2}{2 \tau^2}\right), 1 < k \leq K,
\label{eq_dog}
\end{align}
as the basis functions for constructing $\psi$. The primary motivation to use LET-based activation is its success over the ST in several image deconvolution \cite{pan2013iterative, blu_surelet_deconv} and denoising problems \cite{newsure,blu_surelet, blu_surelet_1}.\\
\indent We take a supervised learning approach to solve the sparse coding problem, with the goal of learning the coefficients $c_k$ optimally for a given training dataset containing measurement-signal pairs. The training objective is to minimize the aggregate error between the predictions made by the network and the corresponding ground-truth signals. ISTA for the sparse estimation problem has a linear rate of convergence \cite{beck2009fast} and performs reasonably well for sufficiently large number of iterations. Therefore, the ST is a reasonable choice for the activation function to begin with, which we intend to improve during the course of training. For this reason, we select the coefficients $c_k$ and the parameter $\tau$ such that $\psi$ approximates the ST function closely. We observed empirically that fixing $\tau=\frac{\nu}{3}$ works well over a large variety of data, and the parameters $c_k$ are computed to minimize the fitting error in a least-squares sense.\\
\indent It is possible to determine the regularizer associated with a sparsity promoting proximal operator. In the context of the LET, we show that there exists a regularizer $g$ for which the LET $\psi$ turns out to be the proximal operator. Before proceeding further, we introduce the following definition of the inverse of the LET, which, depending on the parameters, may not always be invertible in the conventional sense due to lack of strict monotonicity of the LET.
\begin{definition}
The inverse of the LET-based activation function $\psi$ is defined as
\begin{equation}
\label{eq: inverse_let_def}
\psi^{-1} (q) \buildrel\triangle\over = 
\begin{cases}
\max \{r\!: q = \psi(r) \} & \text{\,\,for\,\,} q > 0, \\
0 & \text{\,\,for\,\,}q = 0, \text{\,\,and} \\
\min \{r\!: q = \psi(r) \}, & \text{\,\,for\,\,}  q < 0.
\end{cases}
\end{equation}
\end{definition}
An example of the inverse of an LET activation is shown in Figure~\ref{fig: LET_inv_Reg}(b).
\begin{proposition}
The LET-based activation function $\psi$ is a proximal operator corresponding to the symmetric regularizer $g(x)$ constructed as
\begin{equation*}
\label{eq: regularizer}
 g(x) = 
\begin{cases}
\tilde{g}(x), & \text{\,\,if\,\,}x\geq 0,\\
\tilde{g}(-x), & \text{\,\,otherwise}, \\
\end{cases}
\end{equation*}
where $\tilde{g}(x) =\frac{1}{\nu} \int_{0}^{x}  (\psi^{-1}(q) - q)\, \mathrm{d}q$, for $x\geq 0$.
\end{proposition}
\textit{Proof:}
The function $\tilde{g}$ is well-defined since the integrand is bounded and continuous everywhere except at $q=0$. The function $g$ so constructed is differentiable. Since $g(x)$ is symmetric, it has an anti-symmetric derivative $\frac{ \mathrm{d} g(x)}{\mathrm{d} x} = \frac{1}{\nu}\left(\psi^{-1}\left(x \right) - x\right)$, $x \in \mathbb{R}$. By the definition of the proximal operator in \eqref{prox_op_def}, we have that
\begin{equation*}
\label{eq: let_reg_opt}
P_{\nu}^{g}(u) = x^* \in \arg \underset{x \in \mathbb{R}}{\min}\text{\,\,} \frac{1}{2} |x-u |^2 + \nu \, g(x).
\end{equation*}
It follows from elementary calculus that 
\begin{equation}
x^*-u + \nu \, \left.\frac{ \mathrm{d} g(x)}{\mathrm{d} x}\right|_{x^*}=0.
\label{prox_op_derivation_step1}
\end{equation}
Substituting $\nu\left.\frac{ \mathrm{d} g(x)}{\mathrm{d} x}\right|_{x^*}=\psi^{-1}\left(x^* \right) - x^*$ in \eqref{prox_op_derivation_step1}, we get that $u=\psi^{-1}\left(x^* \right)$, thereby leading to $x^*=\psi(u)$. \hfill $\blacksquare$\\
\indent The regularizers corresponding to the LET-based proximal operators shown in Figure~\ref{sub_fig: LET} are illustrated in Figure~\ref{sub_fig: reg}. We observe that controlled variations in the LET coefficients result in a wide variety of sparsity-encouraging regularizers. The parametric control to design a rich class of sparsity-inducing regularizers is a major advantage of the LET. Note that the definition of the proximal operator does not make any assumption regarding the convexity of $g$. In fact, a classic example of non-convex sparsity encouraging regularizer is the $\ell_0$ quasi-norm, whose proximal operator is the hard-thresholding function. Notably, the $\ell_p$ quasi-norm, for any $0\leq p<1$, is a non-convex regularizer encouraging sparsity. 

\begin{figure*}[t!]
	\centering
	\def\svgwidth{0.9\textwidth}
	\begingroup\makeatletter\def\f@size{8.5}\check@mathfonts
	\def\maketag@@@#1{\hbox{\m@th\large\normalfont#1}}%
	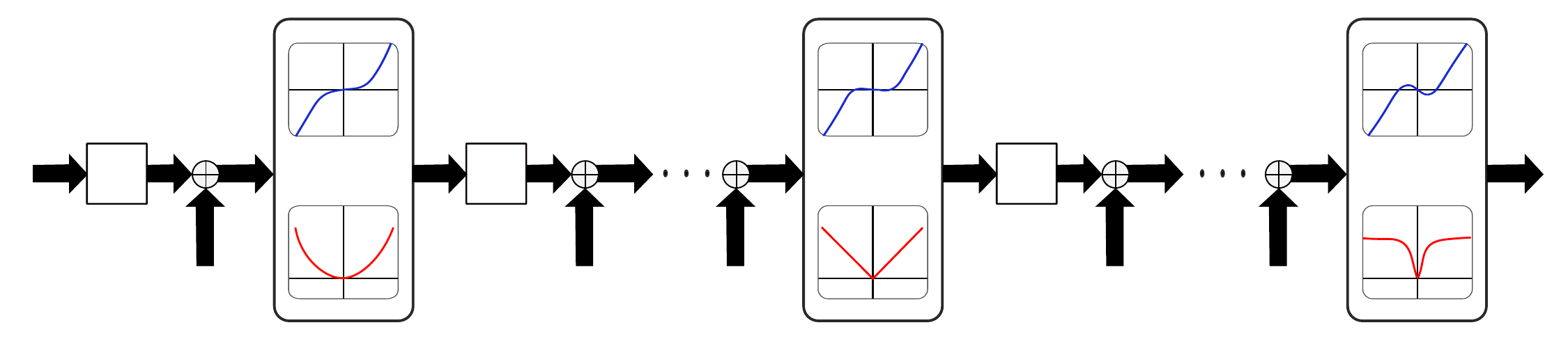
	\endgroup
	\caption{\small(Color online) A schematic of the proposed \textit{LETnetVar} with $L$ layers. Nonlinear LET shrinkage operators $\psi_{\nu}^{t}$ are applied pointwise on $\tilde{\mathbf{x}}^t$ at layer-$t$. Typical examples of LET activation functions and the corresponding regularizers are shown in blue and red, respectively, for each layer.}
	\label{fig: LETnet}
\end{figure*}
\section{Architecture of the Proposed Network}
\label{LETnet_architecture_sec}
\indent For a proximal operator $P_{\nu}^{g}$, with $\nu=\lambda \eta$, the operations in (\ref{eq: Proximal_op_definition}) can be expressed compactly as 
\begin{eqnarray}
\bold x^{t+1} &=& \psi\left(\tilde{\bold x}^{t+1} \right),\label{eq: post_activation}\\
\text{where}\quad \tilde{\bold x}^{t+1}&=&\mathbf{W}\mathbf{x}^t + \mathbf{b},
\label{eq: pre_activation}
\end{eqnarray}
with $\bold W$ and $\bold b$ as defined in Section \ref{sec_prox_grad_explain}. The proximal operator $P_{\nu}^{g}$ in \eqref{eq: Proximal_op_definition} is replaced by the activation $\psi$ applied element-wise on $\tilde{\bold x}^{t+1}$. Each iteration in (\ref{eq: post_activation}) can be viewed as an affine operation followed by a nonlinear activation. This perspective establishes a bridge between iterative algorithms for sparse coding and feed-forward NN.
Motivated by the flexibility offered by LET-based activation and efficient parametric control over the induced regularizers, as elucidated in Section \ref{sec_LET_prop}, we employ them as nonlinear activation functions. In the sequel, we refer to a NN with LET-based activation as \textit{LETnet}. Unlike the classical feed-forward NN, \textit{LETnet} contains fixed weights and biases determined by the sensing matrix $\bold A$ and the measurement $\bold y$ at every layer, whereas the LET coefficients are optimized during training, for which we propose efficient algorithms. We consider two architectures: \textit{LETnetFixed}, where the LET parameters across all the layers are tied; and \textit{LETnetVar}, where the LET parameters are allowed to vary across layers. The numerical experiments indicate that the \textit{LETnetVar} results in better reconstruction compared with \textit{LETnetFixed}, owing to higher flexibility.\\
\indent Recently, Kamilov et al. \cite{7442798} proposed to employ a parametric linear combination of cubic B-splines instead of the soft-threshold, to parametrize the activation function, which is optimized during training and kept fixed across all layers. The resulting network is referred to as minimum mean-squared error ISTA (\textit{MMSE-ISTA}). We shall show that, one of the advantages of using an LET-based activation is that it takes remarkably less number of basis functions ($K\approx 5$) to cover a fairly rich class of regularizers without sacrificing the representational ability of the network. In contrast, \textit{MMSE-ISTA} requires $K\approx 8000$ cubic B-spline bases. Therefore, the proposed NN model is less likely to over-fit a fixed training set, as it uses less number of free parameters.\\
\indent The proposed \textit{LETnetVar} architecture is illustrated in Figure \ref{fig: LETnet}. The initial estimate $\mathbf{x}^0$ (typically an all-zero vector) is passed through the layers of the \textit{LETnetVar} to obtain an estimate $\mathbf{x}^L$ of the true sparse signal. Once the activation function coefficients are optimized for a fixed sensing matrix $\bold A$, it is straightforward to predict $\bold x$ given a new noisy and compressive measurement -- it is simply a matter of computing one forward pass through the network. While the computational complexities of \textit{LETnetFixed}/\textit{LETnetVar} per layer and ISTA per iteration are identical, the network has far fewer layers than the number of iterations of ISTA, leading to an overall lower complexity. The reconstruction accuracy is also superior in case of \textit{LETnetVar}/\textit{LETnetFixed}.
\begin{algorithm}[!t]
	\caption{The \textsc{Back-Propagation} algorithm for the \textit{LETnetVar} architecture}
	\label{algo: backprop}
	\begin{algorithmic}[1]
	\Require {A training pair $\left(\bold y, \bold x\right)$, the sensing matrix $\bold A$}
	\State Perform a forward pass through the \textit{LETnetVar} to compute $\bold x^t$ and $\tilde{\bold x}^t$, $t=1:L$.
		\State \textbf{Initialize} $t \leftarrow L$ and $\nabla_{\bold x^{t}}J\leftarrow \bold x^L-\bold x$. 
		\For{$t = L$ down to $1$}
		\State $\nabla_{\bold c^{t}}J=\left(\boldsymbol \Phi^{t}\right)^\top\nabla_{\bold x^{t}}J$, where $\Phi^t_{i,k}=\phi_k\left( \tilde{x}_i^t\right)$
		\State $\nabla_{\tilde{\bold x}^{t}}J=\text{diag}\left(\psi^{'(t)}\left( \bold x^{t}\right)\right)\nabla_{\bold x^{t}}J$ \vspace{1mm}
		\State $\nabla_{\bold x^{t-1}}J=\bold W^\top\nabla_{\tilde{\bold x}^{t}}J$
		
		\EndFor\vspace{1mm}
		\Ensure {The gradient vectors $\nabla_{\bold c^{t}}J$, for $t=1:L$}
	\end{algorithmic}
\end{algorithm}
\begin{algorithm}[htb]
	\caption{\textsc{Hes\_Vec}: An algorithm for computing the Hessian-vector product $\bold H \bold u$, for any vector $\bold u$, where $\bold H$ is the Hessian matrix of $J$ evaluated at $\bold c$.}
	\label{algo: HP}
	\begin{algorithmic}[1]
		\Require{The vector $\bold u$ and $\bold c$, such that $\bold H=\nabla^2 J(\bold c)$}
		\State Run the \textsc{Back\_Propagation} algorithm and obtain $\bold x^t$, $\tilde{\bold x}^t$, and $\nabla_{\bold x^{t}}J$, for $t=1:L$.
		
		\State \textbf{Forward-pass}: Beginning with $t=1$ and $\mathcal{R}_{\bold u}\left(\bold x^{0} \right)=\bold 0$, where $\mathcal{R}_{\bold u}$ is as defined in \eqref{R_oper_def}, iteratively run through the following three steps until $t\leq L$: 
\begin{enumerate}
\item $\mathcal{R}_{\bold u}\left(\tilde{\mathbf{x}}^t \right)=\bold W \mathcal{R}_{\bold u}\left(\bold x^{t-1} \right)$, 
\item $\mathcal{R}_{\bold u}(\phi_k(\tilde{\mathbf{x}}^t))=\text{diag}\left(\phi_k^{'}\left(\tilde{\bold x}^t\right)\right) \mathcal{R}_{\bold u}\left(\tilde{\mathbf{x}}^t \right)$, for $k = 1:K$, and
\item $\mathcal{R}_{\bold u}\left({\mathbf{x}}^t \right)=\sum_{k=1}^{K} \left[ u^t_k \phi_k(\tilde{\mathbf{x}}^t) + c^t_k\bold M_k \mathcal{R}_{\bold u}\left(\tilde{\mathbf{x}}^t \right)\right]$,
\end{enumerate}
where $\bold M_k=\text{diag}\left(\phi_k^{'}\left(\tilde{\bold x}^t\right)\right)$.		
\State \textbf{Initialize the backward-pass}: Set $t\leftarrow L$, $\mathcal{R}_{\bold u}\left( \nabla_{{\bold x}^{t}}J\right)\leftarrow \mathcal{R}_{\bold u}\left( \bold x^{L}\right)$, which is computed in the forward-pass.
		\For{$t=L$ down to 1} 	
		\State $\mathcal{R}_{\bold u}(\phi_k^{'}(\tilde{x}_j^{t}))=\phi_k^{''}\left(\tilde{x}_j^{t}\right) \mathcal{R}_{\bold u}\left(\tilde{x}_j^{t} \right)$
		\State $\mathcal{R}_{\bold u}\left(\psi^{'(t)}\left(\tilde{x}_j^{t}\right)  \right)=\sum_{k=1}^{K} \left[ u^{t}_k \phi_k^{'}(\tilde{x}_j^{t}) + c^{t}_k \mathcal{R}_{\bold u}(\phi_k^{'}(\tilde{x}_j^{t}))\right]$
		
		\State Evaluate the following using $\Phi^t_{i,k}=\phi_k\left( \tilde{x}_i^t\right)$ and the values of $\mathcal{R}_{\bold u} \left( \boldsymbol \Phi^t\right)$ calculated in Step 2 of the forward-pass:
		\begin{eqnarray*}
\mathcal{R}_{\bold u}\left( \nabla_{\bold c^t}J \right)=\left( \boldsymbol \Phi^t\right)^\top \mathcal{R}_{\bold u}\left( \nabla_{\bold x^t}J \right)+\left(\mathcal{R}_{\bold u} \left( \boldsymbol \Phi^t\right)\right)^\top \nabla_{\bold x^t}J.
\end{eqnarray*}
		
		\State Compute $\mathcal{R}_{\bold u}\left( \nabla_{\tilde{\bold x}^{t}}J\right)$ using \eqref{hv_BPvar_step6_2}.
%
		\State Evaluate $\mathcal{R}_{\bold u}\left(\nabla_{\bold x^{t-1}}J \right)=W^\top \mathcal{R}_{\bold u}\left(\nabla_{\tilde{\bold x}^{t}}J \right)$.
		
		\EndFor
		\Ensure { The Hessian-vector product $\mathbf{H}\bold u $, which is the concatenation of $\mathcal{R}_{\bold u}\left(\nabla_{\bold c^{t}}J\right)$, for $t=1:L$.}	
		\end{algorithmic}
\end{algorithm}
\begin{algorithm}[htb]
	\caption{Hessian-free optimization for minimizing $J(\bold c)$}
	\label{algo: opt_hess_free}
	\begin{algorithmic}[1]
		\For{$i = 1, 2, \cdots$, until convergence,} 
		\State Find $\bold{g}_i =\left. \nabla J(\bold c) \right|_{\bold c= \bold c_i}$ using \textsc{Back-Propagation}. \vspace{1mm}
		\State Find the optimal direction $\boldsymbol \delta_{\bold c}^*$ using the conjugate-gradient algorithm, that uses the {\textsc{Hes\_Vec}} routine for computing Hessian-vector products. 
		\State Perform line-search to minimize $J$ at $\bold c_i$ in the optimal direction $\boldsymbol \delta_{\bold c}^*$ returned by the conjugate-gradient algorithm.  
		\EndFor
	\end{algorithmic}
\end{algorithm}
\subsection{Training \textit{LETnetVar}: Learning Optimal Shrinkage}
\label{sub_LETnet_training}
\indent For a given $\bold A$, the training dataset $\mathcal{D}$ consists of $N$ examples $\{(\mathbf{y}_q, \mathbf{x}_q)\}_{q=1}^{N}$, where $\mathbf{y}_q = \mathbf{Ax}_q + \boldsymbol \xi_q$. The random noise vectors $\boldsymbol \xi_q$ are assumed to be independent and identically distributed. Let $\mathbf{c}^t  \in  \mathbb{R}^K$, $t=1:L$, be the coefficients of the LET activation at layer $t$. For the $q^{\text{th}}$ example in the dataset, the prediction $\mathbf{x}^L_q$ of the $L^{\text{th}}$ layer is a function of the corresponding measurement vector $\mathbf{y}_q$ and the set of LET coefficient vectors $\bold c=\left\{\bold c^t\right\}_{t=1}^{L}$. For convenience, we assume that $\bold c$ is vector of size $KL$, with the $\bold c^t$s stacked on top of each other. The optimal set of activation parameters $\bold c^*$ is obtained by minimizing the squared estimation error 
\begin{align}
\label{eq: opt_for_coefs}
 J(\bold c)= \frac{1}{2}\text{\,}\sum_{q=1}^{N} \| \mathbf{x}^L_q\left(\bold y_q,\bold c\right) - \mathbf{x}_q \|_2^2,
\end{align}
over all training examples. The optimization requires knowledge of the gradient of $J(\bold c)$ with respect to $\bold c$, for which we derive associated back-propagation algorithms (cf. Algorithm \ref{algo: backprop} and Section \ref{backprop} for \textit{LETnetVar}, and supplementary material for \textit{LETnetFixed}). The optimization of $J(\bold c)$ using vanilla GD tends to diverge, unless a very small step size is chosen. The reason for divergence is the exploding gradient \cite{pascanu2012difficulty}, primarily caused by the unboundedness of $\phi_1$ in the LET expansion. We overcome this problem by the \textit{Hessian-Free optimization} (HFO) technique \cite{martens2010deep}. In the $i^{\text{th}}$ epoch of HFO \cite{martens2010deep} (summarized in Algorithm~\ref{algo: opt_hess_free}), the search direction $\boldsymbol \delta_{\bold c}^{*}$ is obtained by minimizing a locally quadratic Taylor-series approximation $J_i^{(q)}\left(\bold c\right)$ to the actual cost $J(\bold c)$ at the $i^{\text{th}}$ iterate $\bold c_i$:
\begin{equation}
\label{eq: quad_approx}
\begin{aligned}[b]
J_i^{(q)}\left(\bold c_i+\boldsymbol \delta_{\bold c}\right)= J\left(\bold c_i \right) +  \boldsymbol \delta_{\bold c}^\top \bold g_i  + \frac{1}{2} \boldsymbol \delta_{\bold c}^\top  \mathbf{H}_i \boldsymbol \delta_{\bold c},
\end{aligned}
\end{equation}
where $ \bold g_i =\left. \nabla J(\bold c) \right|_{\bold c= \bold c_i}$,  $ \mathbf{H}_i = \left. \nabla^2 J(\bold c) \right|_{\bold c=\bold c_i}$, and $\boldsymbol \delta_{\bold c}$ is the search direction to be chosen optimally at every iteration by minimizing a regularized quadratic approximation:
\begin{equation}
\label{eq: opt_srch_dir}
\boldsymbol \delta_{\bold c}^* = \arg \underset{\boldsymbol \delta_{\bold c}}{\min}\text{\,\,} J_i^{(q)}\left(\bold c_i+\boldsymbol \delta_{\bold c}\right) + \gamma \|\boldsymbol \delta_{\bold c}\|_2^2.
\end{equation}
The above formulation is also known as the {\it trust region method}, which is well known in the optimization literature. The regularization term in (\ref{eq: opt_srch_dir}) ensures that the overall Hessian matrix $\left(\mathbf{H}_i +\gamma \bold I\right)$ is positive-definite and well-conditioned. Solving (\ref{eq: opt_srch_dir}) using Newton's method could be computationally demanding, as the matrix $\left(\mathbf{H}_i +\gamma \bold I\right)$, has to be inverted in each iteration. In \textit{LETnetFixed}, this is not a problem as the Hessian is only of size $K \times K$. In \textit{LETnetVar}, this could be a problem since the Hessian now has dimensions $KL \times KL$. The inversion could be avoided by substituting the Newton method with a few steps of the conjugate-gradient (CG) algorithm. In this case, explicit knowledge of $\bold H_i$ is not required, but it is required to compute the Hessian-vector product $\mathbf{H}_i \bold u$, for any vector $\bold u$ \cite{martens2010deep}. This can be done using an additional round of back-propagation (cf. Algorithm 2). Thus, the Hessian-vector product computation is only twice as expensive as the gradient computation. For details of the derivation, please refer to Section \ref{hvp}.\\
\indent An alternative is to approximately compute the Hessian-vector product using finite-difference approximation: 
\begin{equation}
\mathbf{H}_i \bold u =\underset{\epsilon \rightarrow 0}{\lim}\text{\,}\frac{1}{\epsilon}\left(\left. \nabla J\left(\bold c\right)\right|_{\bold c= \bold c_i+\epsilon \bold u}-\left. \nabla J\left(\bold c\right)\right|_{\bold c= \bold c_i}\right),
\label{hess_vec_prod_FD}
\end{equation}
which may be attractive when the dimension of $\bold c$ is small. We employ the finite-difference approximation for \textit{LETnetFixed} since the computational overhead is small.\\
\indent The additional computation involved in HFO, compared with standard GD, is due to the CG step in Algorithm \ref{algo: opt_hess_free}. However, in practice, only a small number of iterations of CG suffice to obtain an accurate estimate of the optimal search direction $\boldsymbol \delta_{\bold c}^*$ \cite{martens2010deep}. In the $i^{\text{th}}$ training epoch, the coefficients are updated as $\bold c_i \leftarrow \bold c_i +  \boldsymbol \delta_{\bold c}^*$ using the optimal direction $\delta_{\bold c}^*$. The trust-region parameter $\gamma$ in \eqref{eq: opt_srch_dir} is chosen using back-tracking. Each internal iteration of CG computes a Hessian-vector product to perform exact line search. HFO is a second-order optimization technique, and offers superior convergence performance than GD, provided that the LET parameters are appropriately initialized. In our implementation, the coefficients $\bold c$ are initialized such that the activation function in each layer closely fits the optimal ST operator, which guarantees that, to start with, the recovery performance is on par with that of ISTA.
\section{Training \textit{LETnetVar}: Derivation of Back-propagation and Hessian-vector product}
\label{derivation_letnetvar}
\subsection{Back-propagation for \textit{LETnetVar}}
\label{backprop}
\indent For brevity of notation, we derive the back-propagation algorithm for computing the gradient of the cost
\begin{align}
\label{eq: cst_one_eg}
J(\mathbf{c}^1, \mathbf{c}^2, \cdots, \mathbf{c}^L) = \frac{1}{2}\left\|\mathbf{x}^L - \mathbf{x}\right\|_2^2,
\end{align}
for one training pair $(\mathbf{y,x})$. The overall gradient is simply an accumulation of the gradients over all training pairs. The partial derivative of the cost $J$ with respect to $c_k^t$, the $k^{\text{th}}$ LET coefficient at layer $t$, is expressed as
\begin{equation}
\frac{\partial J}{\partial c_k^t}=\sum_{i=1}^{n}\frac{\partial J}{\partial x_i^t} \frac{\partial x_i^t}{\partial c_k^t}, t = 1 : L, k=1 : K,
\label{BPvar_step1}
\end{equation}
where $x_i^t$ is the $i^{\text{th}}$ coordinate of $\bold x^t$. Expanding the partial derivative $\frac{\partial J}{\partial x_i^t}$ gives
\begin{equation}
\frac{\partial J}{\partial x_i^t}=\sum_{j=1}^{n}\frac{\partial J}{\partial \tilde{x}_j^{t+1}} \frac{\partial \tilde{x}_j^{t+1}}{\partial x_i^t},
\label{BPvar_step2}
\end{equation}
where $\tilde{\bold x}^{t+1}=\mathbf{W}\mathbf{x}^t + \mathbf{b}$, as defined in \eqref{eq: pre_activation}. Consequently, we have $\frac{\partial \tilde{x}_j^{t+1}}{\partial x_i^t}=W_{ji}$, the $(j,i)^{\text{th}}$ entry of $\mathbf{W}$. To evaluate $\frac{\partial J}{\partial \tilde{x}_j^{t+1}}$, we write
\begin{equation}
\frac{\partial J}{\partial \tilde{x}_j^{t+1}}=\sum_{\ell=1}^{n}\frac{\partial J}{\partial x_{\ell}^{t+1}} \frac{{\partial x_{\ell}^{t+1}}}{\partial \tilde{x}_j^{t+1}}.
\label{BPvar_step3}
\end{equation}
Since the proximal operator is applied coordinate-wise on $\tilde{\bold x}^{t+1}$ to compute ${\bold x}^{t+1}$, $\frac{{\partial x_{\ell}^{t+1}}}{\partial \tilde{x}_j^{t+1}}$ vanishes for $\ell \neq j$, leading to
\begin{equation}
\frac{\partial J}{\partial \tilde{x}_j^{t+1}}=\frac{\partial J}{\partial x_{j}^{t+1}} \frac{{\mathrm{d} x_{j}^{t+1}}}{\mathrm{d}\tilde{x}_j^{t+1}}.
\label{BPvar_step4}
\end{equation}
Using the notation $\psi^{'(t+1)}\left( \bold x^{t+1}\right)$ to indicate the vector whose $j^{\text{th}}$ entry is $\frac{{\mathrm{d} \psi^{t+1}\left(\tilde{x}_j^{t+1} \right)}}{\mathrm{d}\tilde{x}_j^{t+1}}=\frac{{\mathrm{d} x_{j}^{t+1}}}{\mathrm{d}\tilde{x}_j^{t+1}}$, one can write \eqref{BPvar_step4} compactly as
\begin{equation}
\nabla_{\tilde{\bold x}^{t+1}}J=\text{diag}\left(\psi^{'(t+1)}\left( \bold x^{t+1}\right)\right)\nabla_{\bold x^{t+1}}J,
\label{BPvar_step5}
\end{equation}
where $\text{diag}\left(\bold u\right)$ denotes a diagonal matrix with the vector $\bold u$ on its diagonal, and $\nabla_{\bold u}J$ is  the gradient of $J$ with respect to $\bold u$. Further, using $\frac{\partial \tilde{x}_j^{t+1}}{\partial x_i^t}=W_{ji}$, we express \eqref{BPvar_step2} as
\begin{equation}
\nabla_{\bold x^{t}}J=\bold W^\top \nabla_{\tilde{\bold x}^{t+1}}J.
\label{BPvar_step6}
\end{equation}
Finally, by defining an $n\times K$ matrix $\boldsymbol \Phi^t$ whose $\left(i,k\right)^{\text{th}}$ entry is given by $\Phi^t_{i,k}=\frac{\partial x_i^t}{\partial c_k^t}=\phi_k\left( \tilde{x}_i^t\right)$, \eqref{BPvar_step1} is written as
\begin{equation}
\nabla_{\bold c^{t}}J=\left(\boldsymbol \Phi^t\right)^ \top \nabla_{\bold x^{t}}J.
\label{BPvar_step7}
\end{equation}
Therefore, the gradient $\nabla_{\bold c^{t}}J$, of the cost $J$ can be computed recursively as follows:
\begin{enumerate}
\item $\nabla_{\tilde{\bold x}^{t}}J=\text{diag}\left(\psi^{'(t)}\left( \bold x^{t}\right)\right)\nabla_{\bold x^{t}}J$, from \eqref{BPvar_step5},
\item $\nabla_{\bold x^{t-1}}J=\bold W^\top \nabla_{\tilde{\bold x}^{t}}J$ from \eqref{BPvar_step6}, and
\item $\nabla_{\bold c^{t-1}}J=\left(\boldsymbol \Phi^{t-1}\right)^\top \nabla_{\bold x^{t-1}}J$ from \eqref{BPvar_step7},
\end{enumerate}
for $t = L, L-1, \cdots, 2$, starting with the initializations $\nabla_{\bold x^{L}}J=\bold x^L-\bold x$ and $\nabla_{\bold c^L}J=\left(\boldsymbol \Phi^L\right)^\top\nabla_{\bold x^{L}}J$. The recursive gradient computation using the back-propagation algorithm is performed after doing a forward computation through the network in order to evaluate $\bold x^t$ and $\tilde{\bold x}^t$, for  $t=1:L$. 
\subsection{Exact Computation of the Hessian-vector Product}
\label{hvp}
\indent The key to exact computation of the Hessian-vector product lies in the definition of the directional derivative operator \cite{pearlmutter1994fast}. For a vector-valued function $\bold h(\bold c)$ of a vector $\bold c$, the derivative along $\bold u$ is defined as
\begin{align}
\mathcal{R}_{\bold u}(\mathbf{h(c)})= \lim_{\alpha\rightarrow 0} \frac{\mathbf{h}(\mathbf{c}+ \alpha\mathbf{u}) - \mathbf{h}(\mathbf{c})}{\alpha} = \left. \frac{\mathrm{d}\mathbf{h}(\mathbf{c}+\alpha\mathbf{u})}{\mathrm{d} \alpha} \right|_{\alpha=0}.
\label{R_oper_def}
\end{align}
The last equality effectively shows that $\mathcal{R}_{\bold u}$ behaves like a standard derivative operator and is therefore linear. The dimensions of $\mathcal{R}_{\bold u}(\mathbf{h(c)})$ are same as that of $\mathbf{h}$. The properties of $\mathcal{R}_{\bold u}$ have been established in \cite{pearlmutter1994fast}. For the sake of completeness, we recall the key properties below:
\begin{enumerate}
\item $\mathcal{R}_{\mathbf{u}}(\beta\mathbf{h(c)}) =\beta \mathcal{R}_{\mathbf{u}}(\mathbf{h(c)})$, for any scalar $\beta$;
\item $\mathcal{R}_{\mathbf{u}}(\mathbf{c}) = \mathbf{u}$, for any $\bold u$;
\item $\mathcal{R}_{\mathbf{u}}(\mathbf{c}_0)=\bold{0}$, where ${\bold c}_0$ is a constant vector;
\item $\mathcal{R}_{\mathbf{u}}(\mathbf{h}_1(\bold c) + \bold h_2(\bold c)) = \mathcal{R}_{\mathbf{u}}(\mathbf{h}_1(\bold c)) +\mathcal{R}_{\mathbf{u}}(\mathbf{h}_2(\bold c))$, for two vector-valued functions $\bold h_1$ and $\bold h_2$;
\item $\mathcal{R}_{\bold u}$ applied on the inner product of $\bold h_1$ and $\bold h_2$ satisfies the product rule: $\displaystyle \mathcal{R}_{\mathbf{u}}\left(\mathbf{h}_1(\bold c)^\top \bold h_2(\bold c)\right) =  \mathcal{R}_{\mathbf{u}}(\mathbf{h}_1(\bold c))^\top \mathbf{h}_2(\bold c)
\hspace{-0.02in} + \mathbf{h}_1(\bold c)^\top \mathcal{R}_{\mathbf{u}}(\mathbf{h}_2(\bold c))$; and
\item $\mathcal{R}_{\mathbf{u}}(\bold h_1 \left(\bold h_2(\bold c)\right)) = \frac{\partial \bold h_1}{\partial \mathbf{h}_2}  \mathcal{R}_{\mathbf{u}}(\bold h_2(\bold c))$, where $ \frac{\partial \bold h_1}{\partial \mathbf{h}_2}$ is an $m_1\times m_2$ Jacobian, $m_1$ and $m_2$ being the dimensions of $\bold h_1$ and $\bold h_2$, respectively. The $(i,j)^{\text{th}}$ entry of  $\frac{\partial \bold h_1}{\partial \mathbf{h}_2}$ is $\left.\frac{\partial \bold h_1}{\partial \mathbf{h}_2}\right|_{ij}=\frac{\partial \bold h_{1i}}{\partial \mathbf{h}_{2j}}$, for $i = 1 : m_1$ and $j = 1 : m_2$.
\end{enumerate}
We use $\mathcal{R}_{\bold u}$ to compute the Hessian-vector product in the $i^{\text{th}}$ training epoch. For a vector $\bold u$ having the same dimension as $\bold c$, the Hessian-vector product is defined as
\begin{equation}
\mathbf{H}_i \bold u = \lim_{\alpha \rightarrow 0} \frac{
	\left. \nabla_{\bold c} J(\bold c) \right|_{\bold c=\bold c_i +\alpha \bold u} - 
	\left. \nabla_{\bold c} J(\bold c) \right|_{\bold c=\bold c_i}
}{\alpha}.
\label{R_oper_def1}
\end{equation}
Equivalently,
\begin{equation}
\mathbf{H}_i \bold u = \left. \mathcal{R}_{\bold u} \left( \nabla_{\bold c} J(\bold c) \right) \right|_{\bold c = \bold c_i},
\label{R_oper_def2}
\end{equation}
where $\nabla_{\bold c} J(\bold c)$ is a vector of dimension  $KL\times 1$, constructed by stacking the vectors $\nabla_{\bold c^{t}}J$, for $t = 1 : L$. Hereafter, we drop the training epoch index $i$ for notational brevity. To compute the Hessian-vector product, the $\mathcal{R}_{\bold u}$ operator has to be applied on the vectors $\nabla_{\bold c^{t}}J$, as suggested by \eqref{R_oper_def2}. In the following, we derive a back-propagation algorithm to compute $\mathcal{R}_{\bold u}\left(\nabla_{\bold c^{t}}J\right)$. Analogous to the back-propagation algorithm derived in Section~\ref{backprop}, we require $\mathcal{R}_{\bold u}(\bold x^t)$ and $\mathcal{R}_{\bold u}(\bold {\tilde x}^t)$ for computations in the backward pass. These required quantities are computed in the forward pass as explained in the following.

\subsubsection{Forward Pass}
Observe that
	\begin{eqnarray}
	\label{eq: R_forward}
	\mathcal{R}_{\bold u}(\mathbf{x}^t) &=& \mathcal{R}_{\bold u} \left(\sum_{k=1}^{K} c^t_k \phi_k(\tilde{\mathbf{x}}^t) \right), \nonumber \\
	&\stackrel{(a)}{=}& \sum_{k=1}^{K} \left( \mathcal{R}_{\bold u}(c^t_k) \phi_k(\tilde{\mathbf{x}}^t) + c^t_k \mathcal{R}_{\bold u}(\phi_k(\tilde{\mathbf{x}}^t)) \right),  \nonumber\\
	&\stackrel{(b)}{=} &\sum_{k=1}^{K} \left[ u^t_k \phi_k(\tilde{\mathbf{x}}^t) + c^t_k \mathcal{R}_{\bold u}(\phi_k(\tilde{\mathbf{x}}^t))\right], 
	\end{eqnarray}
\normalsize
where $(a)$ is obtained by applying Properties 1, 4, and 5; and (b) by applying Property 2. Using Property 6, $\mathcal{R}_{\bold u}(\phi_k(\tilde{\mathbf{x}}^t))$ evaluates to
\begin{equation}
\mathcal{R}_{\bold u}(\phi_k(\tilde{\mathbf{x}}^t))=\bold M_k \mathcal{R}_{\bold u}\left(\tilde{\mathbf{x}}^t \right),
\end{equation}
where the Jacobian $M_{k(i, j)}=\frac{\partial \phi_k(\tilde{x}_i^t)}{\partial \tilde{x}_j^t}$, for $1\leq i,j \leq n$. Clearly, $M_{k(ij)}$ vanishes for $i\neq j$, thereby leading to 
\begin{equation}
\mathcal{R}_{\bold u}(\phi_k(\tilde{\mathbf{x}}^t))=\text{diag}\left(\phi_k^{'}\left(\tilde{\bold x}^t\right)\right) \mathcal{R}_{\bold u}\left(\tilde{\mathbf{x}}^t \right),
\label{old_eq31}
\end{equation}
where $\phi_k^{'}\left(\tilde{\bold x}^t\right)$ is an $n$-dimensional vector whose $i^{\text{th}}$ entry is given by $\frac{\partial \phi_k(\tilde{x}_i^t)}{\partial \tilde{x}_i^t}$. Finally, $\mathcal{R}_{\bold u}\left(\tilde{\mathbf{x}}^t \right)$ can be written as 
\begin{equation}
\mathcal{R}_{\bold u}\left(\tilde{\mathbf{x}}^t \right)=\mathcal{R}_{\bold u}\left(\bold W \bold x^{t-1}+\bold b \right)=\bold W \mathcal{R}_{\bold u}\left(\bold x^{t-1} \right),
\end{equation}
using Properties 1, 3, and 4. The quantities $\mathcal{R}_{\bold u}\left(\tilde{\mathbf{x}}^t \right)$ and $\mathcal{R}_{\bold u}\left(\mathbf{x}^t \right)$ are computed in the forward pass as follows, starting with $t=1$ until $t = L$:
\begin{enumerate}
\item $\mathcal{R}_{\bold u}\left(\tilde{\mathbf{x}}^t \right)=\bold W \mathcal{R}_{\bold u}\left(\bold x^{t-1} \right)$;
\item $\mathcal{R}_{\bold u}(\phi_k(\tilde{\mathbf{x}}^t))=\text{diag}\left(\phi_k^{'}\left(\tilde{\bold x}^t\right)\right) \mathcal{R}_{\bold u}\left(\tilde{\mathbf{x}}^t \right)$, for $k = 1:K$; and
\item $\mathcal{R}_{\bold u}\left({\mathbf{x}}^t \right)=\sum_{k=1}^{K} \left[ u^t_k \phi_k(\tilde{\mathbf{x}}^t) + c^t_k\mathcal{R}_{\bold u}(\phi_k(\tilde{\mathbf{x}}^t))\right]$;
\end{enumerate}
where $\mathcal{R}_{\bold u}\left(\bold x^{0} \right)=\bold 0$, as the initialization $\bold x^{0}$ is independent of $\bold c$.
\subsubsection{Backward Pass}
Applying $\mathcal{R}_{\bold u}$ on both sides of \eqref{BPvar_step1} and using Property 5 of the operator $\mathcal{R}_{\bold u}$, we get
\begin{eqnarray}
\mathcal{R}_{\bold u}\left(\frac{\partial J}{\partial c_k^t}\right)&=&\sum_{i=1}^{n}\mathcal{R}_{\bold u}\left(\frac{\partial J}{\partial x_i^t}\right) \frac{\partial x_i^t}{\partial c_k^t}+\frac{\partial J}{\partial x_i^t}\mathcal{R}_{\bold u}\left(\frac{\partial x_i^t}{\partial c_k^t}\right),\nonumber\\
&=&\sum_{i=1}^{n}\mathcal{R}_{\bold u}\left(\frac{\partial J}{\partial x_i^t}\right) \phi_k\left( \tilde{x}_i^t\right)+\frac{\partial J}{\partial x_i^t}\mathcal{R}_{\bold u}\left(\phi_k\left( \tilde{x}_i^t\right)\right),\nonumber\\
\label{hv_BPvar_step1}
\end{eqnarray}
using the fact that $\frac{\partial x_i^t}{\partial c_k^t}= \phi_k\left(\tilde{x}_i^t\right)$. Consolidating over $k = 1 : K$, and using matrix-vector notation, \eqref{hv_BPvar_step1} is written compactly as
\begin{eqnarray}
\mathcal{R}_{\bold u}\left( \nabla_{\bold c^t}J \right)=\left( \boldsymbol \Phi^t\right)^\top \mathcal{R}_{\bold u}\left( \nabla_{\bold x^t}J \right)+\left(\mathcal{R}_{\bold u} \left( \boldsymbol \Phi^t\right)\right)^\top \nabla_{\bold x^t}J,
\label{hv_BPvar_step1_vec}
\end{eqnarray}
where $\Phi^t_{i,k}=\phi_k\left( \tilde{x}_i^t\right)$ and $\mathcal{R}_{\bold u} \left( \boldsymbol \Phi^t\right)$ is calculated using \eqref{old_eq31}. Further, applying $\mathcal{R}_{\bold u}$ on both sides of \eqref{BPvar_step2}, we obtain
\begin{equation*}
\mathcal{R}_{\bold u}\left(\frac{\partial J}{\partial x_i^t}\right)=\sum_{j=1}^{n}\mathcal{R}_{\bold u}\left(\frac{\partial J}{\partial \tilde{x}_j^{t+1}}\right) \frac{\partial \tilde{x}_j^{t+1}}{\partial x_i^t}+\frac{\partial J}{\partial \tilde{x}_j^{t+1}}\mathcal{R}_{\bold u}\left(\frac{\partial \tilde{x}_j^{t+1}}{\partial x_i^t}\right),
\label{hv_BPvar_step2}
\end{equation*}
which simplifies to
\begin{equation}
\mathcal{R}_{\bold u}\left(\frac{\partial J}{\partial x_i^t}\right)=\sum_{j=1}^{n}\mathcal{R}_{\bold u}\left(\frac{\partial J}{\partial \tilde{x}_j^{t+1}}\right) W_{ji},
\label{hv_BPvar_step3}
\end{equation}
since $\frac{\partial \tilde{x}_j^{t+1}}{\partial x_i^t}=W_{ji}$ and $\mathcal{R}_{\bold u}\left(W_{ji}\right)=0$. The vectorized representation of \eqref{hv_BPvar_step3} is given by
\begin{equation}
\mathcal{R}_{\bold u}\left(\nabla_{\bold x^t}J \right)={\bold W}^\top \mathcal{R}_{\bold u}\left(\nabla_{\tilde{\bold x}^{t+1}}J \right).
\label{hv_BPvar_step3_vec}
\end{equation}
Finally, applying the $\mathcal{R}_{\bold u}$ operator on both sides of \eqref{BPvar_step4} gives
\begin{equation*}
\mathcal{R}_{\bold u}\left(\frac{\partial J}{\partial \tilde{x}_j^{t+1}}\right)=\mathcal{R}_{\bold u}\left(\frac{\partial J}{\partial x_{j}^{t+1}}\right) \frac{{\mathrm{d} x_{j}^{t+1}}}{\mathrm{d}\tilde{x}_j^{t+1}}+\frac{\partial J}{\partial x_{j}^{t+1}} \mathcal{R}_{\bold u}\left(\frac{{\mathrm{d} x_{j}^{t+1}}}{\mathrm{d}\tilde{x}_j^{t+1}}  \right).
\label{hv_BPvar_step4}
\end{equation*}
Substituting $\frac{{\mathrm{d} x_{j}^{t+1}}}{\mathrm{d}\tilde{x}_j^{t+1}} =\psi^{'(t+1)}\left(\tilde{x}_j^{t+1}\right)$ leads to
\begin{eqnarray}
\mathcal{R}_{\bold u}\left(\frac{\partial J}{\partial \tilde{x}_j^{t+1}}\right)&=&\mathcal{R}_{\bold u}\left(\frac{\partial J}{\partial x_{j}^{t+1}}\right) \psi^{'(t+1)}\left(\tilde{x}_j^{t+1}\right)\nonumber\\&+&\frac{\partial J}{\partial x_{j}^{t+1}} \mathcal{R}_{\bold u}\left(\psi^{'(t+1)}\left(\tilde{x}_j^{t+1}\right)  \right).
\label{hv_BPvar_step5}
\end{eqnarray}
Expanding $\mathcal{R}_{\bold u}\left(\psi^{'(t+1)}\left(\tilde{x}_j^{t+1}\right)  \right)$ using Property 6, we obtain that
\begin{eqnarray}
&&\mathcal{R}_{\bold u}\left(\psi^{'(t+1)}\left(\tilde{x}_j^{t+1}\right)  \right)=\mathcal{R}_{\bold u}\left( \sum_{k=1}^{K} c^{t+1}_k \phi_k^{'}(\tilde{x}_{j}^{t+1})\right)\nonumber\\
&=&\sum_{k=1}^{K} \left[ u^{t+1}_k \phi_k^{'}(\tilde{x}_j^{t+1}) + c^{t+1}_k \mathcal{R}_{\bold u}(\phi_k^{'}(\tilde{x}_j^{t+1}))\right].
\label{hv_BPvar_step6_1}
\end{eqnarray}
Following Property 6, we get
\begin{equation}
\mathcal{R}_{\bold u}(\phi_k^{'}(\tilde{x}_j^{t+1}))=\phi_k^{''}\left(\tilde{x}_j^{t+1}\right) \mathcal{R}_{\bold u}\left(\tilde{x}_j^{t+1} \right),
\label{hv_BPvar_step6}
\end{equation}
using similar arguments that we used to obtain \eqref{old_eq31}.
Thus, it is possible to compute the vector $\mathcal{R}_{\bold u}\left(\psi^{'(t+1)}\left(\tilde{\bold x}^{t+1}\right)  \right)$ of dimension $n\times 1$ following \eqref{hv_BPvar_step6_1}, \eqref{hv_BPvar_step6}, and using $\mathcal{R}_{\bold u}\left(\tilde{\bold x}^{t+1} \right)$ evaluated in the forward pass, enabling us to compute the vectorized representation of \eqref{hv_BPvar_step5} as
\small
\begin{eqnarray}
\mathcal{R}_{\bold u}\left( \nabla_{\tilde{\bold x}^{t+1}}J\right)&=&\text{diag}\left(\psi^{'(t+1)}\left(\tilde{\bold x}^{t+1}\right)\right)\mathcal{R}_{\bold u}\left( \nabla_{{\bold x}^{t+1}}J\right)\nonumber\\
&+&\text{diag}\left(\mathcal{R}_{\bold u}\left(\psi^{'(t+1)}\left(\tilde{\bold x}^{t+1}\right)  \right) \right)\nabla_{{\bold x}^{t+1}}J.\nonumber\\
\label{hv_BPvar_step6_2}
\end{eqnarray}
\normalsize
For initialization of the backward pass, we observe that $\mathcal{R}_{\bold u}\left( \nabla_{{\bold x}^{L}}J\right)=\mathcal{R}_{\bold u}\left( \bold x^L-\bold x\right)=\mathcal{R}_{\bold u}\left(\bold x^L\right)$, which is saved during the forward computation. The back-propagation algorithm for computing the Hessian-vector product is summarized in Algorithm \ref{algo: HP}, where we use $\mathcal{R}_{\bold u}\left( \bold x^{t}\right)$ and $\mathcal{R}_{\bold u}\left( \tilde{\bold x}^{t}\right)$ stored in the forward pass and $\nabla_{{\bold x}^{t}}J$ computed in Step 5 of Algorithm \ref{algo: backprop}, for $t=1:L$.
\begin{figure*}[t!]
			\centering
			\captionsetup[subfigure]{labelformat=empty}
			\begin{subfigure}{0.33\textwidth}
				\centering
				\includegraphics[width=2.33in]{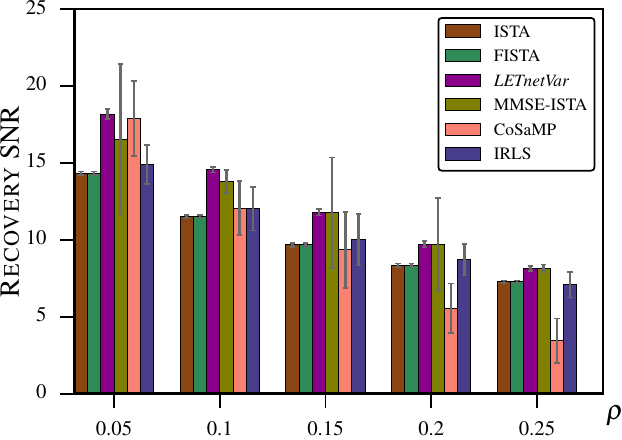}
				\caption{(a) $\text{SNR}_{\text{input}}=10$ dB}
				\label{sub_fig: snr_10}
			\end{subfigure}%
			\hspace{-2.5mm}
			\begin{subfigure}{0.33\textwidth}
				\centering
				\includegraphics[width=2.33in]{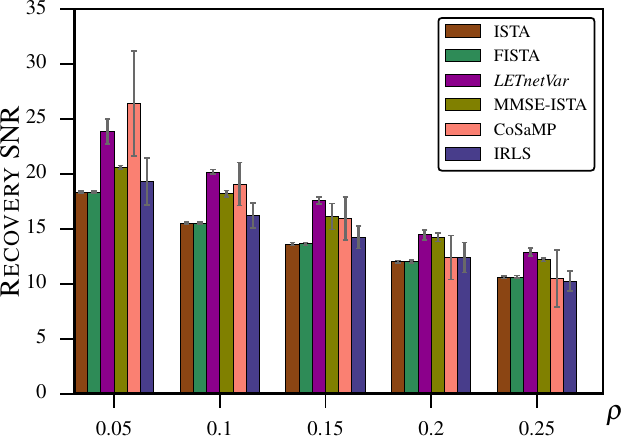}
				\caption{(b) $\text{SNR}_{\text{input}}=15$ dB}
				\label{sub_fig: snr_15}
			\end{subfigure} 
			\begin{subfigure}{0.33\textwidth}
				\centering
				\includegraphics[width=2.33in]{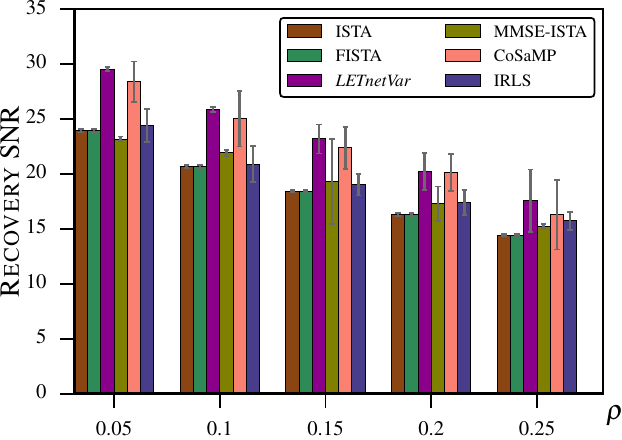}
				\caption{(c) $\text{SNR}_{\text{input}}=20$ dB}
				\label{sub_fig: snr_20}
			\end{subfigure} \\
			\begin{subfigure}{0.33\textwidth}
				\centering
				\includegraphics[width=2.33in]{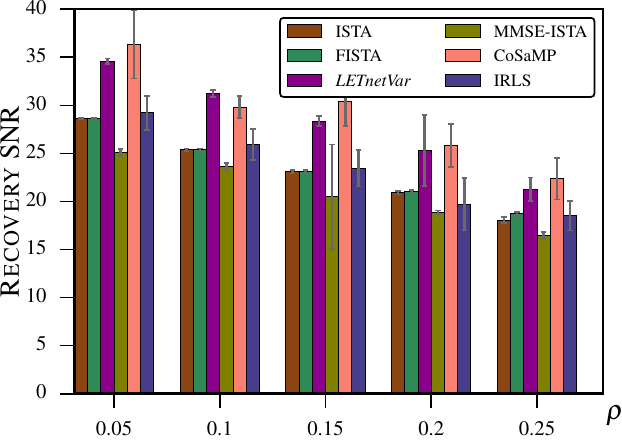}
				\caption{(d) $\text{SNR}_{\text{input}}=25$ dB}
				\label{sub_fig: snr_25}
			\end{subfigure}
			\begin{subfigure}{0.33\textwidth}
				\centering
				\includegraphics[width=2.33in]{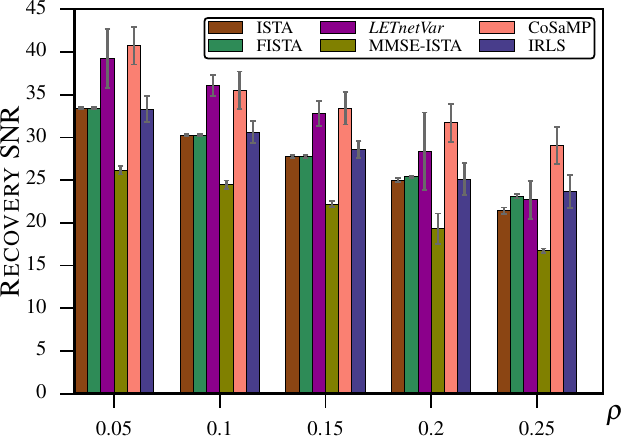}
				\caption{(e) $\text{SNR}_{\text{input}}=30$ dB}
				\label{sub_fig: snr_30}
			\end{subfigure}
			\begin{subfigure}{0.33\textwidth}
			\centering
			\includegraphics[width=2.33in]{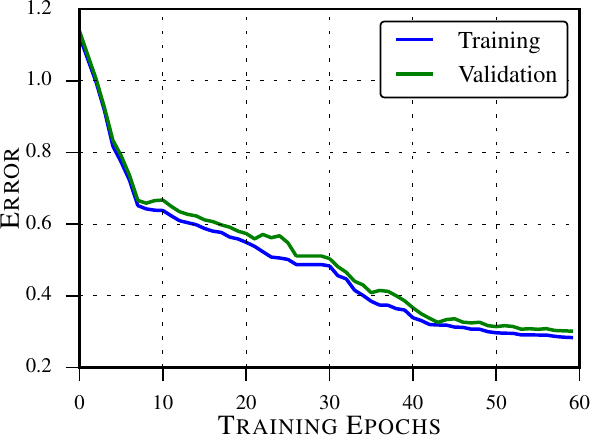}
			\caption{\; \; \; (f) Training and validation error}
			\label{sub_fig: LETnet_training}
		\end{subfigure}
			\caption{\small (Color online) The variation of the (ensemble averaged) reconstruction SNR for different algorithms as a function of the sparsity parameter $\rho$, corresponding to different measurement/input SNRs. 
				\hfill}
			\label{fig: exp_LETnet}
		\end{figure*}
\section{Numerical Validation of \textit{LETnetVar}}
\label{num_valid_letnet_sec}
\indent We next validate the recovery performance of \textit{LETnetVar} on synthesized signals. The comparison of \textit{LETnetFixed} with \textit{LETnetVar} will be reported in Section \ref{sec_exp_validation_fLETnet}.
\subsection{Experimental Setting}
\label{sec_exp_settings}
\indent We consider the reconstruction of a sparse signal $\bold x$ of dimension $n = 256$ from $m = \left \lceil{0.7n}\right \rceil$-dimensional noisy compressive measurement $\bold y=\bold A \bold x +\boldsymbol \xi$. 
The entries of $\bold A$ are independent and follow $\mathcal{N}(0,1/m)$ (Gaussian distribution). The sparse signal $\bold x$ is generated as $\bold x = \bold x_{\text{supp}}\odot\bold x_{\text{mag}}$, where the operator $\odot$ denotes the element-wise product. The entries of $\bold x_{\text{supp}}$ are either $0$ or $1$, drawn independently from a Bernoulli distribution with parameter $\rho \in [0,1]$. Increasing the parameter $\rho$ results in a denser signal $\bold x$. The elements of $\bold x_{\text{mag}}$ follow $\mathcal{N}(0,1)$. The training dataset used for learning the nonlinearities of \textit{LETnetVar} and \textit{MMSE-ISTA} consists of $N=100$ examples. We conduct experiments for different values of the sparsity-controlling parameter $\rho$ and the input SNR: $\text{SNR}_{\text{input}}=\frac{\left\| \bold A \bold x \right\|_2}{\left\| \boldsymbol \xi \right\|_2}$. The reconstruction SNR during testing, obtained using different algorithms, is calculated by averaging over $10$ independent trials, where, in each trial, $N_{\text{test}}=100$ test signals are considered.\\  
\indent For performance comparison, we consider five state-of-the-art sparse recovery algorithms: (i) ISTA, which employs the ST proximal operator; (ii) FISTA, which is an accelerated version of ISTA; (ii) CoSaMP, which is a greedy algorithm; (iv) the iteratively reweighted least-squares (IRLS) algorithm \cite{irls_daubechies}, which is based on majorization-minimization; and (v) \textit{MMSE-ISTA}, which is a training-based algorithm developed by Kamilov and Mansour \cite{7442798}. The optimum regularizer $\lambda_{\text{opt}}$ for ISTA, FISTA, and IRLS is chosen from $10$ logarithmically spaced points in the interval $[10^{-5}, 0.1]$ such that the reconstruction SNR during training is maximized. An all-zero vector is used as the initialization for all algorithms to ensure fairness in comparison. The CoSaMP algorithm is supplied with the exact knowledge of the number of nonzero entries in $\bold x$.
	\begin{figure*}[t!]		
		\def\svgwidth{\textwidth}
		\begingroup\makeatletter\def\f@size{9}\check@mathfonts
		\def\maketag@@@#1{\hbox{\m@th\large\normalfont#1}}%
		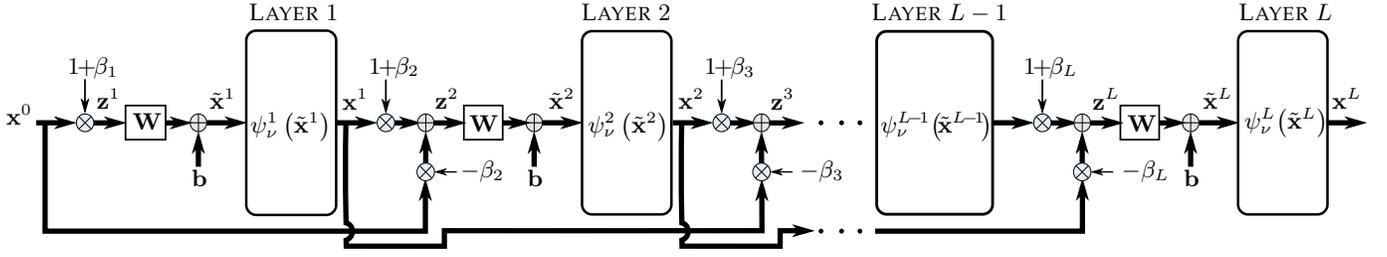
		\endgroup
		\caption{\small Architecture of \textit{fLETnetVar}. Every layer (except the first layer) is connected to two of its previous layers. The activation functions are untied across layers.}
		\label{fig: f_LETnet}
	\end{figure*}
\subsection{Training Details of \textit{LETnetVar} and \textit{MMSE-ISTA}}
\label{sec_training_details_LETnet}
\indent For a fixed sensing matrix $\mathbf{A}$, a training dataset consisting of $N=100$ examples is generated for learning the optimal nonlinearities of \textit{LETnetVar} and \textit{MMSE-ISTA}. A validation set containing $20$ examples is used during training in order to avoid over-fitting. The trained network is tested on a dataset containing $N_{\text{test}}=100$ examples drawn according to the same statistical model that was used to generate the training dataset (cf. Section \ref{sec_exp_settings}). The procedure is repeated $T=10$ times with different datasets, corresponding to $10$ different sensing matrices, to measure the ensemble-averaged performance of \textit{LETnetVar} and \textit{MMSE-ISTA}. The optimal regularization parameter $\lambda$ for ISTA is chosen as explained in Section \ref{sec_exp_settings} and the same value is used in \textit{MMSE-ISTA}, as suggested in \cite{7442798}. The parameter $\lambda_{\text{opt}}$ used in training \textit{LETnetVar} is chosen by cross-validation over five values of $\lambda$ placed logarithmically in the interval $[0.05, 0.5]$. Both networks contain $L=100$ layers.\\
\indent For the \textit{MMSE-ISTA} architecture, $K=501$ coefficients are used to parametrize the activation function using cubic B-splines, and the same set of coefficients are shared across all the layers. The \textit{MMSE-ISTA} network is trained using GD with a step-size of $\eta=10^{-4}$, as suggested in \cite{7442798}, and the algorithm is terminated when the number of training epochs exceeds $1000$.\\ 
\indent The LET-based activation function in \textit{LETnetVar} is parametrized with $K=5$ coefficients in every layer, leading to a total of $500$ parameters, which is comparable to the number of free parameters in \textit{MMSE-ISTA}. The HFO algorithm for training \textit{LETnetVar} is terminated after $60$ epochs.

\subsubsection{Initialization and termination criteria for CG}
\label{sec_CG_algo}
\indent The CG algorithm required to compute the optimal direction $\boldsymbol \delta_{\bold c}^*$ in every epoch of HFO is initialized with that obtained in the previous epoch of HFO. Theoretically, it takes $KL$ iterations for the CG algorithm to compute an exact solution to \eqref{eq: opt_srch_dir}, since the variable $\bold c$ is of dimension $KL$. Martens et al. \cite{martens2010deep} reported that it suffices to run the CG algorithm in each HFO iteration only a few times according to a preset criterion, for example, a criterion based on the relative improvement defined as  
\begin{equation*}
\theta \stackrel{\Delta}{=} \frac{Q\left(\boldsymbol \delta_{\bold c}^{\left(i_2\right)}\right) - Q\left(\boldsymbol \delta_{\bold c}^{\left(i_2-i_1\right)}\right)}{Q\left(\boldsymbol \delta_{\bold c}^{\left(i_2\right)}\right)},
\end{equation*}
where $Q\left(\boldsymbol \delta_{\bold c}\right)=J_i^{(q)}\left(\bold c_i+\boldsymbol \delta_{\bold c}\right) + \gamma \|\boldsymbol \delta_{\bold c}\|_2^2$, for $0<i_1<i_2$. The CG iterations are terminated when $|\theta| < i_1\epsilon$, for $\epsilon>0$ is a user-specified parameter. We employ the same stopping criterion and set $i_1=\text{max}(10, 0.1i_2)$ following \cite{martens2010deep}.

 As the training objective $J \left(\bold c \right)$ is non-convex and is optimized using CG with a regularized quadratic approximation, a backtracking step is introduced after the end of the CG loop to determine the optimal direction $\boldsymbol{\delta^*_c}$. To determine the \textit{trust-region}, the regularization parameter $\gamma$ in (\ref{eq: opt_srch_dir}) is varied using the Levenberg-Marquardt approach, wherein one computes the reduction ratio $\mathit{r}$, defined as
\begin{align*}
\mathit{r} \stackrel{\Delta}{=} \frac{J(\mathbf{c}_i+\boldsymbol \delta_{\bold c}^*)- J(\mathbf{c}_i)}{J^{(q)}_i(\mathbf{c}_i+\boldsymbol \delta_{\bold c}^*)- J^{(q)}_i(\mathbf{c}_i)},
\end{align*}
in each epoch of the HFO and updates $\gamma$ as 
\begin{equation*}
\gamma \leftarrow
\begin{cases}
\frac{2}{3}\gamma & \text{\,\,if\,\,}r< \frac{1}{4},\text{\,\,and}\\
\frac{3}{2}\gamma & \text{\,\,if\,\,}r> \frac{3}{4}.
\end{cases}
\end{equation*}
As a result, the trust-region of the quadratic approximation used in CG is contracted (enlarged) if the reduction ratio $r$ is small (large).
\subsection{Explanation and Interpretation of the Result}
\label{sec_letnet_results_explanation}
\indent The average and standard deviation of the reconstruction SNR,
computed over $10$ independent trials, as a function of $\rho$ and input SNR, for different methods is shown in Figure~\ref{fig: exp_LETnet}. We observe that \textit{LETnetVar} outperforms ISTA and FISTA by a margin of about $4$ dB, while exhibiting relatively small deviation, and by a margin of $3$ dB over IRLS. The CoSaMP algorithm, which is by far the best greedy algorithm for sparse recovery, results in $1$ to $2$ dB higher reconstruction SNR than \textit{LETnetVar}, particularly for smaller values of $\rho$ and higher input SNR. However, unlike ISTA, FISTA, \textit{LETnet}, and IRLS, one needs to know the sparsity level of the ground-truth to execute CoSaMP successfully. The improvement obtained using \textit{LETnetVar} over \textit{MMSE-ISTA} algorithm is largely due to accurate and parsimonious modeling of the activation function using LET. An illustration of how the error of \textit{LETnetVar} varies as a function of training epochs, over the training and validation set, is shown in Figure~\ref{sub_fig: LETnet_training}. The training error profile shows that  \textit{LETnetVar} does not \textit{underfit} the data, particularly because the vanishing gradient problem has been avoided by the use of HFO. The validation error profile does not show any signs of \textit{overfitting} for the number of epochs considered. If the number of epochs is increased further, the validation error might saturate.\\
\indent \textit{LETnetVar} took about $60$ training epochs for learning the parameters, which is significantly smaller than the number required to learn a typical DNN. The reduction in the number of training epochs and faster convergence are due to two factors: (i) the specific initialization of $\bold c$, which ensures that the network outputs an estimate that is at least as good as that of ISTA; and (ii) HFO, which is an efficient way of performing second-order optimization. In contrast, \textit{MMSE-ISTA} requires significantly more (approximately $1000$) training epochs, as it uses GD for training.  

\begin{figure*}[!t]		

	\captionsetup[subfigure]{labelformat=empty}
	\begin{subfigure}{0.33\textwidth}
		\centering
		\includegraphics[width=2.33in]{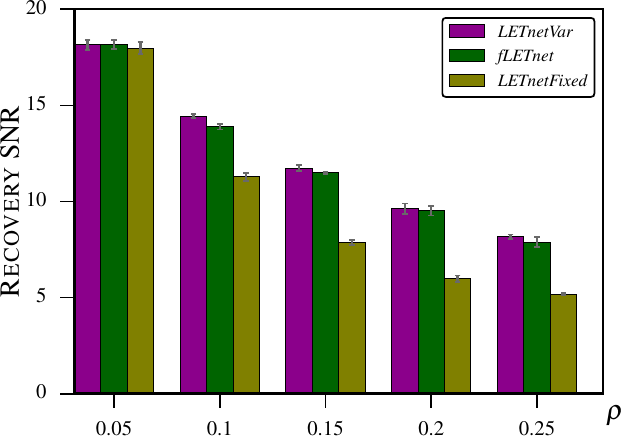}
		\caption{\; \; \; (a) $\text{SNR}_{\text{input}}=10$ dB}
		\label{sub_fig: f_snr_10}
	\end{subfigure}%
	\begin{subfigure}{0.33\textwidth}
		\centering
		\includegraphics[width=2.33in]{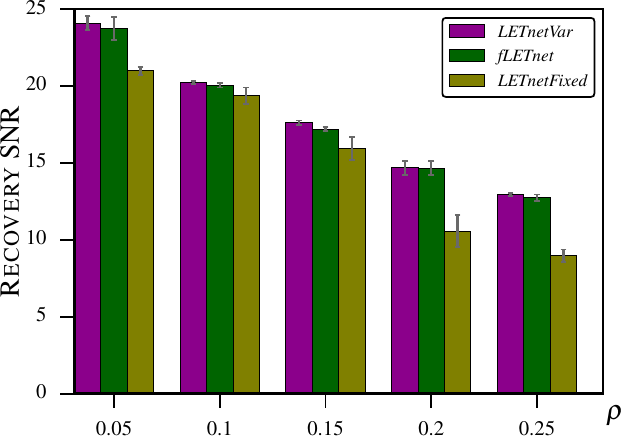}
		\caption{\; \; \; (b) $\text{SNR}_{\text{input}}=15$ dB}
		\label{sub_fig: f_snr_15}
	\end{subfigure} 
	\begin{subfigure}{0.33\textwidth}
		\centering
		\includegraphics[width=2.33in]{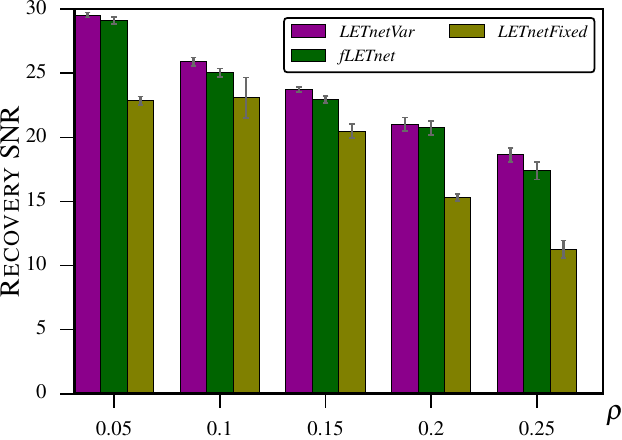}
		\caption{\; \; \; (c) $\text{SNR}_{\text{input}}=20$ dB}
		\label{sub_fig: f_snr_20}
	\end{subfigure} \\
	\begin{subfigure}{0.33\textwidth}
		\centering
		\includegraphics[width=2.33in]{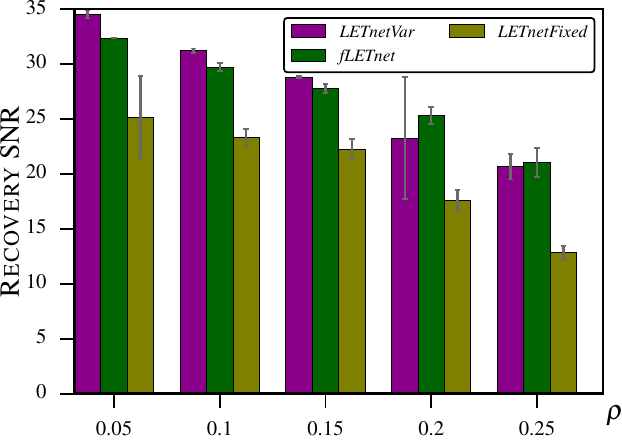}
		\caption{\; \; \; (d) $\text{SNR}_{\text{input}}=25$ dB}
		\label{sub_fig: f_snr_25}
	\end{subfigure}
	\begin{subfigure}{0.33\textwidth}
		\centering
		\includegraphics[width=2.33in]{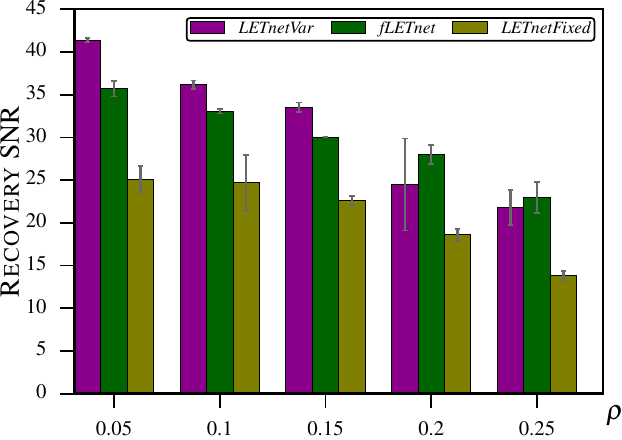}
		\caption{\; \; \; (e) $\text{SNR}_{\text{input}}=30$ dB}
		\label{sub_fig: f_snr_30}
	\end{subfigure}
	\begin{subfigure}{0.33\textwidth}
		\centering
		\includegraphics[width=2.33in]{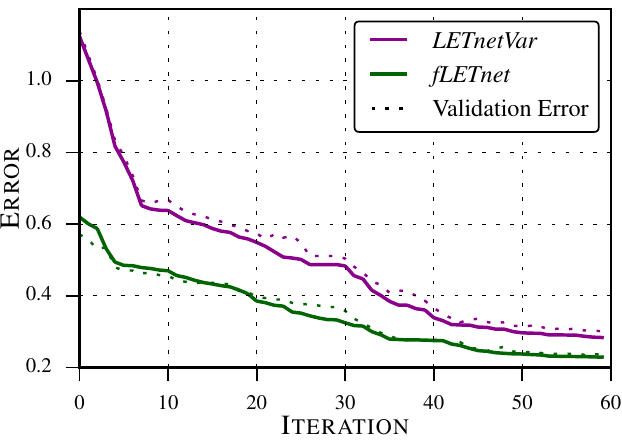}
		\caption{\; \; \; (f) Training and validation error}
		\label{sub_fig: fLETnet_training}
	\end{subfigure}
	\caption{\small (Color online) Comparison of ensemble-averaged reconstruction SNR and its standard-deviation for \textit{fLETnet}, \textit{LETnetVar}, and \textit{LETnetFixed} for different $\rho$ and measurement SNR. The training and the validation errors, as a function of training epochs, are also shown.}
	
	\label{fig: exp_fLETnet}
\end{figure*}
\section{\textit{fLETnet}: An Improved Architecture Motivated by Accelerated Gradient-Descent}
\label{fLETnet_architecture_sec_intro}
\indent The ISTA has a convergence rate of $\mathcal{O}\left(\frac{1}{t}\right)$, where $t$ denotes the iteration index. However, Beck and Teboulle \cite{beck2009fast} have devised a strategy to improve the convergence rate from linear to quadratic, without significantly increasing the computations in each iteration. Their algorithm, referred to as fast ISTA or FISTA, relies on an original idea by Nesterov \cite{nesterov1}, who showed how to accelerate the projected gradient-descent by incorporating a \textit{momentum} factor. The estimate $\bold x^{t+1}$ in the $\left(t+1\right)^{\text{st}}$ iteration of FISTA is computed as 
\begin{equation}
	\label{eq: FISTA}
	\mathbf{x}^{t+1} = T_{\nu} \left(\mathbf{z}^{t+1} - \eta\nabla f\left(\mathbf{z}^{t+1}\right) \right),
\end{equation}
where $T_{\nu}$ is the ST operator and
\begin{align}
	\label{eq: acceleration_step}
	\mathbf{z}^{t+1} = \mathbf{x}^t + \frac{\alpha_t - 1}{\alpha_{t+1}}\left(\mathbf{x}^t - \mathbf{x}^{t-1}\right),
\end{align}
with $\alpha_t$ being the acceleration factor. Choosing $\alpha_{t+1} = \frac{1 + \sqrt{1 + 4 \alpha_t^2}}{2}$, with the initialization $\alpha_1 = 1$ leads to $\mathcal{O}\left(\frac{1}{t^2}\right)$ convergence. As a consequence, FISTA yields an accurate estimate of the ground-truth in remarkably less number of iterations than ISTA.\\
\indent It is natural to expect that a network model inspired by FISTA would also produce accurate estimates with considerably less number of layers than \textit{LETnetVar} or \textit{LETnetFixed}. We leverage the concept of Nesterov's accelerated gradient to construct a new network architecture, which we refer to as the \textit{fLETnet}, to achieve a similar estimation performance as \textit{LETnetVar} with significantly less number of layers.
\subsection{Architecture of the \textit{fLETnet}}
\label{fLETnet_architecture_sec}
\indent Let	 $\beta_{t+1} = \frac{\alpha_t - 1}{\alpha_{t+1}}$; upon rearranging (\ref{eq: acceleration_step}), we get
	\begin{align*}
		\mathbf{z}^{t+1} = \left(1+\beta_{t+1}\right) \mathbf{x}^t - \beta_{t+1} \mathbf{x}^{t-1}.
	\end{align*}
	Furthermore, (\ref{eq: FISTA}) is rewritten as
	\begin{equation}
		\mathbf{x}^{t+1} = \psi^{(t+1)}\left(\mathbf{W}\mathbf{z}^{t+1} + \mathbf{b}\right),
		\label{eq_flet_net1}
	\end{equation}
where $\mathbf{W}$ and $\mathbf{b}$ are as defined in Section \ref{sec_prox_grad_explain}. Similar to \textit{LETnetVar}, we replace the ST-based activation function in \eqref{eq_flet_net1} with a parametrized linear combination of elementary thresholding functions, namely the \textit{LET}s, to construct \textit{fLETnet}. The LET coefficients are allowed to vary across the layers to further increase the flexibility and expressive power of the network. The forward computation through the \textit{fLETnet} is summarized in the following steps: 
\begin{enumerate}
\item $\mathbf{z}^{t} = (1+\beta_t) \mathbf{x}^{t-1} - \beta_t \mathbf{x}^{t-2}$, 
\item $\tilde{\mathbf{x}}^{t} = \mathbf{W}\mathbf{z}^t + \mathbf{b}$, and
\item $\mathbf{x}^t = \psi^{(t)}\left(\tilde{\mathbf{x}}^t\right)$, for $t = 1, 2, \cdots, L$,
\end{enumerate}
starting with the initialization $\mathbf{x}^0 = \mathbf{x}^{-1}=\mathbf{0}$. A schematic of the \textit{fLETnet} architecture is shown in Figure \ref{fig: f_LETnet}. Every layer in \textit{fLETnet} is connected to two previous layers unlike a standard feed-forward NN. The architecture also bears a striking resemblance to \textit{deep residual networks} (DRNs) proposed by He et al. \cite{he2015deep}, who demonstrated significant performance improvements in various computer vision tasks using DRNs. He et al. also showed that the issue of vanishing/exploding gradients for training parameters closer to the first layer can be prevented by establishing connections to the previous layers, so that the network can be trained using GD, even without an accurate initialization. Experimental results in Section~\ref{sec_exp_validation_fLETnet} suggest that the \textit{fLETnet} architecture produces reasonably accurate sparse estimates while containing significantly smaller number of layers than \textit{LETnetVar}, thereby reducing the number of parameters to learn.\\
\indent The cost function for training \textit{fLETnet} is the same as that of \textit{LETnetVar} (cf. Section \ref{sub_LETnet_training}), namely the reconstruction error over the training set. Again, the LET coefficients are initialized such that the activation function approximates the ST operator, so that the initial reconstruction is comparable to that of FISTA. We train the network using HFO to achieve faster convergence. A derivation of the back-propagation algorithm for computing the gradient of the training cost and the Hessian-vector product for the \textit{fLETnet} architecture are given in the supplementary material.   
\begin{figure*}[t]		
			\centering

	\begin{subfigure}{\textwidth}
		\centering
		\includegraphics[width=\textwidth]{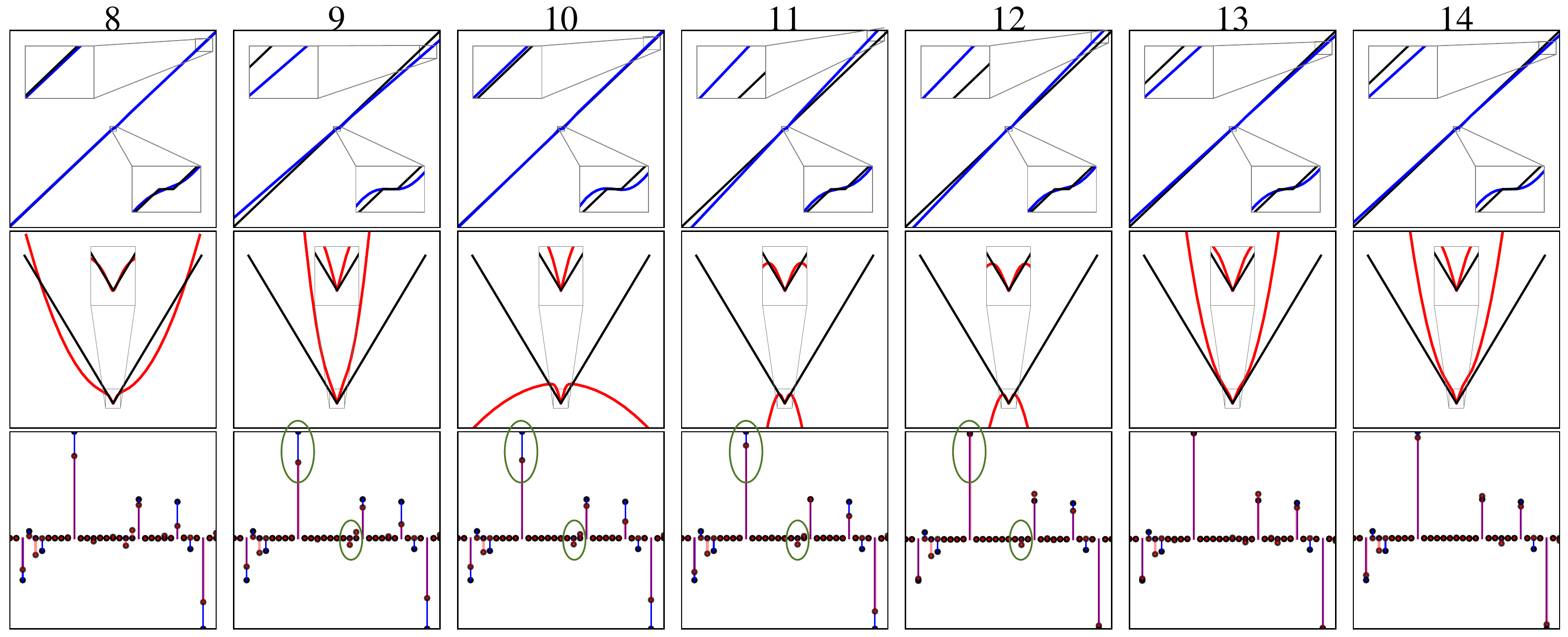}
	\end{subfigure} \\[2mm]
	\caption{The learnt parametric LET functions (blue) of \textit{fLETnet} over a typical dataset and their corresponding induced regularizers (red). The activation functions and the regularizers are compared with ST and the $\ell_1$ penalty (black), respectively. Zoomed-in portions of the estimated signals at the layers are also shown for visual comparison.}
	\label{fig: Learnt_regularizers}
\end{figure*}
\begin{center}

\begin{table}[t]
\centering
\begin{tabular}{|c |c| }
\hline
Algorithm & Per iteration/layer run-time \\
&  (in milliseconds) \\
\hline
ISTA & $0.0331$  \\

\hline
FISTA &$0.0394$   \\

\hline
\textit{LETnetVar}/\textit{LETnetFixed} &$0.0895$   \\

\hline

\textit{fLETnet} &$0.1088$   \\

\hline

\textit{MMSE-ISTA} &$0.6184$   \\

\hline
CoSaMP &$11.7672$   \\

\hline
IRLS &$5.2784$   \\

\hline

\end{tabular}
\caption{\small{A comparison of run-time complexity per iteration/layer, corresponding to various algorithms.}}
\label{table_run_time} 
\end{table}
\end{center}
\subsection{Experimental Validation of \textit{fLETnet}}
\label{sec_exp_validation_fLETnet}
\indent The data generation procedure is the same as that described in Section \ref{sec_exp_settings}. The number of layers in \textit{fLETnet} is taken as $L=50$. A comparison of the reconstruction SNR of \textit{fLETnet} with \textit{LETnetVar} and \textit{LETnetFixed}, for different $\rho$ and input SNR, is shown in Figure~\ref{fig: exp_fLETnet}. We observe that, despite having half the number of layers, \textit{fLETnet} achieves comparable or better performance than the competing architectures, especially when the measurement SNR is low and $\rho$ is high (denser signal). We observe from Figure \ref{sub_fig: fLETnet_training} that the training and validation errors for \textit{fLETnet} reduce at a faster rate than \textit{LETnetVar}. This behavior is attributed to the DRN-type architecture of \textit{fLETnet}, which avoids vanishing/exploding gradients. Furthermore, as expected, \textit{LETnetVar} results in significantly improved performance over \textit{LETnetFixed} (by as much as $5$ dB in some cases and about $2$ dB on the average).\\
\indent The per iteration (per layer for \textit{LETnetVar}, \textit{LETnetFixed}, \textit{fLETnet}, and \textit{MMSE-ISTA}) run-times of various algorithms are reported in Table \ref{table_run_time}. The algorithms are implemented in Python, running on a computer having an intel core i7 $4^{\text{th}}$ generation processor and  8 GB of RAM. We observe that the run-times for \textit{fLETnet}, \textit{LETnetVar}, and \textit{LETnetFixed} are on par with that of ISTA and FISTA. However, the proposed DNN architectures require far less number of layers than the number of iterations in ISTA and FISTA, thereby resulting in less computation time overall. The \textit{MMSE-ISTA} algorithm requires relatively more time than the proposed architectures, mainly because it uses significantly more parameters to model the activation function. The run-times for IRLS and CoSaMP are considerably higher, as they require to compute matrix inversion and pseudo-inverse, respectively, within every iteration.

\subsection{What Regularizers Does the \textit{fLETnet} Learn?}
\indent To gain insights into the functioning of \textit{fLETnet}, it is instructive to study the activation functions and the corresponding regularizers in each layer at the end of training. In Figure~\ref{fig: Learnt_regularizers}, we show an illustration of the typical learnt activations and the regularizers over each layer from 8 to 14 of a trained \textit{fLETnet}. A segment of the corresponding output signals are also shown. The activation functions in all $50$ layers are provided in the supplementary document. Take for instance, layer 8, where the regularizer is closer to the $\ell_1$ penalty over the dynamic range of the input. On the other hand, the induced regularizers in layers 9, 13, and 14 are between the $\ell_1$ and $\ell_2$ penalties. Interestingly, the regularizers in layers 10, 11, and 12 decrease beyond a certain range of inputs and even go negative, thereby encouraging an already large input to increase further in magnitude. Such a behavior is in stark contrast with the familiar $\ell_0/\ell_1$-norm-based regularizers, which never amplify the input. This behavior of the network seems to have a high impact on support recovery. To understand further, let us examine the LET-based activation defined in \eqref{eq_let_activation_def}, for $\left| v\right|>v_{0}$, where $v_0$ is a large constant (for example, $v_0=3\tau$). As the exponential terms in the expression of $\psi$ decay as $\left|v\right|$ increases, we can write $\psi(v) \approx c_1v$ for $\left| v\right|>v_{0}$. As the coefficients are initialized to approximate the ST operator, we have $c_1<1$ prior to training, because the ST output never exceeds its input in magnitude. However, in the process of training, $c_1$ might get updated to a value larger than one, thereby resulting in $\left|\psi(v)\right| \approx c_1\left|v\right|>\left|v\right|$, for $\left|v\right|>v_0$. This is reflected in the corresponding regularizer, as it slopes downwards and even becomes negative for inputs of large magnitudes. However, the induced regularizers across all layers offer a positive penalty for small magnitudes resulting in noise rejection. The network effects two simultaneous operations -- reduction of small amplitudes (noise), and enhancement of large amplitudes, which are due to the signal. The signal component that gets eliminated in one layer in the process of canceling noise can again be recovered in a subsequent layer due to the negative penalty as highlighted in the third row of Figure~\ref{fig: Learnt_regularizers}. Therefore, the regularizers learnt by \textit{fLETnet} cover a wide range between hard and soft thresholding, and even go beyond these, leading to a balance between noise rejection and signal preservation. 
\subsection{Support Recovery in \textit{fLETnet}}
\begin{figure}[t]
	\centering
	\includegraphics[width=\columnwidth]{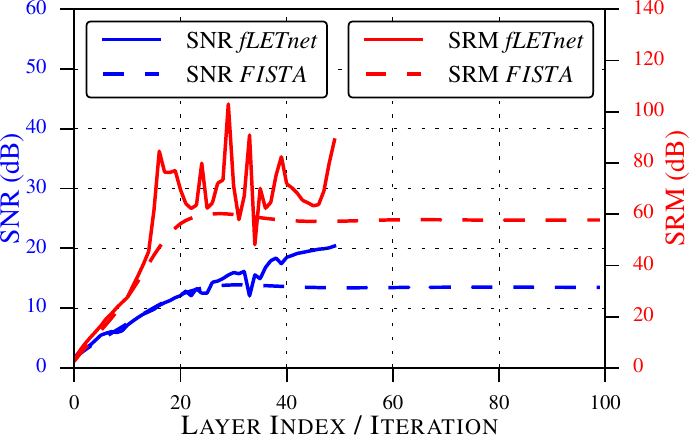}
	\caption{\small (Color online) SNR values (in dB) for intermediate solutions of the FISTA and activations of a trained \textit{fLETnet}. Each SNR value is obtained by averaging over $100$ test examples. Since the \textit{fLETnet} has $50$ layers, the SNR values are shown as a function of the layers up to $50$ only. On the other hand, FISTA was executed for $100$ iterations.}
	\label{fig: FISTA_vs_fLETnet}
\end{figure}

\indent In Figure~\ref{fig: FISTA_vs_fLETnet}, we plot the layer-wise reconstruction SNR, defined as $\text{SNR}_t=\frac{\left\|\bold x^t-\bold x\right\|_2^2}{\left\|\bold x\right\|_2^2}$ on test signals, versus the layer index $t$ of a trained \textit{fLETnet}. The input SNR and the sparsity parameter $\rho$ are taken as $20$ dB and $0.2$, respectively. The variation of reconstruction SNR with respect to the corresponding number of iterations of FISTA is also shown to facilitate comparison. We observe that the recovery SNR for FISTA increases initially and saturates as the number of iterations exceeds $40$, resulting in no further improvement in estimation. However, the evolution of $\text{SNR}_t$ for a trained \textit{fLETnet} exhibits an overall increasing trend, in spite of the local variations. The estimate produced by the final layer is significantly better than the FISTA output, by almost $5.5$ dB.\\
\indent The $\text{SNR}_t$ has a steep increase for the first few layers (up to $20$), nearly the same behavior as that of FISTA. After that, the $\text{SNR}_t$ seems to oscillate before reaching a value about $5.5$ dB higher than FISTA. We conjecture that the network actually stabilizes its support estimate in this regime. To justify further, we consider a {\it support recovery metric} (SRM) based on the evolution of the ratio of the signal energy over the support to that outside of it:
\begin{equation}
\text{SRM}_t = \frac{\|{\mathcal{P}_s(\bold x^t)}\|_2^2}{\|\bold x^t\|_2^2-\|{\mathcal{P}_s(\bold x^t)}\|_2^2},
\end{equation}
where $s$ is the sparsity level, which is assumed to be known solely for the sake of computing the metric; and $\mathcal{P}_s$ denotes the best $s$-sparse approximation operator, which retains only the top $s$ significant entries of its argument.\\
\indent The SRM values, averaged over $100$ realizations, are also shown in Figure~\ref{fig: FISTA_vs_fLETnet}. Upon juxtaposing with the SNR plot, we observe that the transition region in $\text{SNR}_t$ corresponds to a transition to a high value in $\text{SRM}_t$ as well. The region over which support stabilization takes place might depend on the depth of the network. Once the support is estimated, the network is effectively performing a regression task for estimating the amplitudes.
\section{Conclusions and Outlook}
\label{sec_conclusion}
\indent We have addressed the problem of efficient estimation of sparse signals from incomplete noisy measurements, which is of importance in several contemporary signal processing and machine learning applications. We have developed a specialized DNN architecture that enables sparse estimation by unfolding the iterations of soft-thresholding algorithms such as ISTA and FISTA with precomputed weights and biases shared across the layers. In particular, the nonlinear activation is modeled using a combination of threshold functions with an approximation to the soft-threshold as the initialization. Training over the exemplars, equipped with a customized back-propagation algorithm, ensures that the performance is only better than that of ISTA and FISTA. This learning paradigm has the advantage that the DNN is endowed with the capability to discover an ensemble of regularizers, one per layer, that have an overall sparse recovery performance better than $\ell_1$ regularization techniques. The optimization is carried out using an efficient second-order algorithm that does not require explicit computation of the Hessian, which also considerably reduces the number of training epochs. The proposed deep architecture is capable of producing sparser estimates, leading to an overall gain of $4$ to $5$ dB over the state-of-the-art sparse recovery algorithms. The improvement comes at a modest cost of learning only five parameters per layer. If we tie the parameters across layers, the drop in SNR is about $2$ dB, but still above the competing techniques by $2$ to $3$ dB. The proposed LET model for the activation may not always lead to a convex regularizer and associated gains in convergence rate. However, it is known in the literature that convexity of the  regularizer is not necessary \cite{selesnick} for promoting sparsity and may even be a restriction. Viewing the proposed methodology purely from a DNN perspective, we found that there are strong similarities with the recently developed DRNs, which have been shown to outperform conventional feed-forward networks. Another key aspect of the proposed architectures is that the number of parameters to learn is independent of the underlying signal dimension.\\
\indent While the proposed learning paradigm has certain advantages over the standard sparse recovery algorithms, it has certain caveats as well. For example, in a standard CS formulation, exact knowledge of the sensing matrix $\bold A$ is assumed during both sensing and sparse recovery. For an altogether different sensing matrix, the sparse recovery algorithm could simply use the new matrix for reconstruction. On the other hand, in a learning paradigm such as this one, merely switching the matrices will not work.  The network has to be trained all over again with the new matrix in order to achieve optimal reconstruction performance.\\
\indent The MMSE-ISTA formulation, which motivated this work, could also benefit in terms of speed-up during training by employing a second-order method, and the Hessian-vector products could be computed efficiently as done in this paper. The DRN link with the \textit{fLETnet} requires further investigation in terms of optimizing the architecture and parameters.

\section*{Acknowledgments}
The authors would like to thank Ulugbek Kamilov from Mitsubishi Electric Research Laboratory for clarifying several aspects regarding the implementation of the \textit{MMSE-ISTA} algorithm, and Thierry Blu, Chinese University of Hong Kong, for his insights on the LET representation and feedback on the manuscript.

\ifCLASSOPTIONcompsoc
%

\ifCLASSOPTIONcaptionsoff
  \newpage
\fi

\bibliographystyle{IEEEtran}
\bibliography{references}

\section*{Supplemental Material}
The supplementary document contains the following:
\begin{enumerate}
\item Derivation of the back-propagation algorithms for \textit{LETnetFixed} (Section~\ref{supp1}) and \textit{fLETnet} (Section~\ref{supp2});
\item Exact Hessian-vector computation for \textit{fLETnet} (Section~\ref{supp3}); and
\item A depiction of the learnt activations and regularizers across the layers of a trained \textit{fLETnet} (Section~\ref{supp4}).
\end{enumerate}
The equations in the main manuscript are qualified by a prefix `M', whenever they are referred to, to avoid conflict with the equation numbers in this supplementary document.
\section{Back-propagation Algorithm for \textit{LETnetFixed}}
\label{supp1}
\indent The derivation of the back-propagation algorithm for \textit{LETnetFixed} is exactly same as that of \textit{MMSE-ISTA} [28], except that choosing the derivatives-of-Gaussian (DoG) as the basis functions does not restrict the space of induced shrinkages to be shift-invariant. Nonetheless, to make the exposition self-contained, we reproduce the derivation. In \textit{LETnetFixed}, the LET parameters $\bold c=\left\{c_k\right\}_{k=1}^{K}$ are shared across the layers, and are learnt by minimizing the training objective $J(\bold c)= \frac{1}{2}\text{\,}\sum_{q=1}^{N} \| \mathbf{x}^L_q\left(\bold y_q,\bold c\right) - \mathbf{x}_q \|_2^2$. For notational convenience, the index $p$ is dropped while deriving the back-propagation algorithm for gradient computation. Notably, the output $\bold x^t$, for each layer $t$ of \textit{LETnetFixed}, depends on $\bold c$. Differentiating the training error $J\left(\bold c\right)$ with respect to $c_k$, we get
\begin{equation}
\frac{\partial J}{\partial c_k}=\sum_{i=1}^{n}\left({x}_i^L - {x}_i\right)\frac{\partial {x}_i^L}{\partial c_k}.
\label{letnet_fixed_eq1}
\end{equation}
Therefore, one needs to evaluate $\frac{\partial {x}_i^L}{\partial c_k}$, where $i=1:n$ and $k=1: K$, in order to compute the gradient $\nabla_{\bold c}J$. For every layer $t$, $t=1:L$, we have
\begin{equation}
x^{t}_i = \sum_{j=1}^{K}c_j\phi_j\left(\tilde{x}^{t}_i \right), i=1:n.
\label{letnet_fixed_eq2}
\end{equation}
Differentiating both sides of \eqref{letnet_fixed_eq2} with respect to $c_k$, we obtain the following:
\small
\begin{equation}
\frac{\partial x^{t}_i }{\partial c_k}=\phi_k\left(\tilde{x}^{t}_i \right)+\underbrace{\sum_{j=1}^{K}c_j\phi_j'\left(\tilde{x}^{t}_i \right)}_{\psi'\left(\tilde{x}^{t}_i \right)}\frac{\partial \tilde{x}^{t}_i }{\partial c_k}=\phi_k\left(\tilde{x}^{t}_i \right)+\psi'\left(\tilde{x}^{t}_i \right)\frac{\partial \tilde{x}^{t}_i }{\partial c_k}.
\label{letnet_fixed_eq3}
\end{equation}
\normalsize
The term $\displaystyle\frac{\partial \tilde{x}^{t}_i }{\partial c_k}$ is evaluated considering $\tilde{x}^{t}_i = \displaystyle\sum_{\ell=1}^{n}W_{i\ell}x^{t-1}_{\ell}+b_i$. Consequently, \eqref{letnet_fixed_eq3} becomes
\begin{equation*}
\frac{\partial x^{t}_i }{\partial c_k}=\phi_k\left(\tilde{x}^{t}_i \right)+\psi'\left(\tilde{x}^{t}_i \right)\sum_{\ell=1}^{n}W_{i\ell}\frac{\partial {x}^{t-1}_{\ell} }{\partial c_k}.
\label{letnet_fixed_eq4}
\end{equation*}
Using the notation $\Phi^t_{i,k}=\phi_k\left( \tilde{x}_i^t\right)$, we write, for any vector $\bold r\in \mathbb{R}^n$, that
\begin{eqnarray}
\sum_{i=1}^{n}\frac{\partial x^{t}_i }{\partial c_k}r_i&=&\sum_{i=1}^{n}\Phi^t_{i,k}r_i+\sum_{i=1}^{n}\sum_{\ell=1}^{n}\psi'\left(\tilde{x}^{t}_i \right)W_{i\ell}\frac{\partial {x}^{t-1}_{\ell} }{\partial c_k}r_i\nonumber\\
&=&\sum_{i=1}^{n}\Phi^t_{i,k}r_i+\sum_{\ell=1}^{n}\frac{\partial {x}^{t-1}_{\ell} }{\partial c_k}\sum_{i=1}^{n}W_{i\ell}\psi'\left(\tilde{x}^{t}_i \right)r_i.\nonumber\\
\label{letnet_fixed_eq5}
\end{eqnarray}
\normalsize
Let $\frac{\partial \bold x^{t}}{\partial \bold c}$ be the $n\times K$ Jacobian matrix whose $(i,k)^{\text{th}}$ entry is given by $\frac{\partial x^{t}_i }{\partial c_k}$. Then, \eqref{letnet_fixed_eq5} can be expressed compactly using matrix notations as
\small
\begin{equation}
\left(\frac{\partial \bold x^{t}}{\partial \bold c}\right)^\top \bold r=\left(\boldsymbol \Phi^t\right)^\top \bold r+\left(\frac{\partial \bold x^{t-1}}{\partial \bold c}\right)^\top \bold W^\top \text{diag}\left(\psi'\left(\tilde{\bold x}^{t} \right)\right)\bold r.
\label{letnet_fixed_eq5_matrix}
\end{equation}
\normalsize
Writing \eqref{letnet_fixed_eq1} using matrix notation, we observe that the gradient $\nabla_{\bold c}J$ is given by
\begin{equation}
\nabla_{\bold c}J=\left(\frac{\partial \bold x^{L}}{\partial \bold c}\right)^\top \left(\bold x^L-\bold x\right).
\label{letnet_fixed_eq1_matrix}
\end{equation}
Setting $\bold r=\bold r^L=\bold x^L-\bold x$ in \eqref{letnet_fixed_eq5_matrix}, we get
\small
\begin{equation}
\nabla_{\bold c}J=\left(\boldsymbol \Phi^L\right)^\top \bold r^L+\left(\frac{\partial \bold x^{L-1}}{\partial \bold c}\right)^\top \bold W^\top \text{diag}\left(\psi'\left(\tilde{\bold x}^{L} \right)\right)\bold r^L.
\label{letnet_fixed_eq6}
\end{equation}
\normalsize
The back-propagation algorithm is derived by computing the right-hand side of \eqref{letnet_fixed_eq6} recursively, and using the fact that $\frac{\partial \bold x^{0}}{\partial \bold c}=\bold 0$, which is an all-zero matrix of size $n \times K$. The steps of the algorithm are summarized in the following:\\
1) Run a forward pass through \textit{LETnetFixed} to compute ${\bold x}^t$ and $\tilde{\bold x}^t$, for $t=1:L$.\\
2) Initialize $t\leftarrow L$, $\bold r^t \leftarrow \bold x^t-\bold x$, and $\bold g^t \leftarrow \bold 0$. Repeat the following steps until $t=0$, and return $\bold g^0$, which is the required gradient $\nabla_{\bold c}J$, when $t=0$:
\begin{itemize}
\item $\bold g^{t-1}=\bold g^t+\left(\boldsymbol \Phi^t\right)^\top \bold r^t$,
\item $\bold r^{t-1}=\bold W^\top \text{diag}\left(\psi'\left(\tilde{\bold x}^{t} \right)\right)\bold r^t$, and
\item $t\leftarrow t-1$.
\end{itemize}
As the number of parameters to optimize for in the \textit{LETnetFixed} architecture is only $K=5$, the finite difference formula in (M-18) gives an accurate approximation of the Hessian-vector product. Hence, for simplicity, we implement the HFO-based training for \textit{LETnetFixed} using the approximation.

\section{Back-propagation for \textit{fLETnet}}
\label{supp2}
The parameters of \textit{fLETnet} are learnt by optimizing the training error $J(\bold c)= \frac{1}{2}\text{\,}\sum_{q=1}^{N} \| \mathbf{x}^L_q\left(\bold y_q,\bold c\right) - \mathbf{x}_q \|_2^2$, similar to what is done for \textit{LETnetVar} and \textit{LETnetFixed}. For notational brevity, we drop the index $q$ and derive the gradient only for a single training example. The overall gradient is obtained by adding the gradients over the individual training examples.\\    
\indent The gradient vector $\nabla_{\mathbf{c}} J$ is constructed by stacking $\left\{\nabla_{\mathbf{c}^t}J\right\}_{t=1}^{L}$. From (M-26), we have that
\begin{align}
\label{eq: dJ_dct_value}
	\nabla_{\mathbf{c}^t}J = (\mathbf{\Phi}^t)^\top \nabla_{\mathbf{x}^t}J,
	\end{align}
where an $\boldsymbol \Phi^t$ is an $n\times K$ matrix, whose $\left(i,k\right)^{\text{th}}$ entry is given by $\Phi^t_{i,k}=\frac{\partial x_i^t}{\partial c_k^t}=\phi_k\left( \tilde{x}_i^t\right)$. Using the law of the total derivative, we express $\frac{\partial J}{\partial x_i^t}$, the $i^{\text{th}}$ entry of $\nabla_{\mathbf{x}^t}J$, for $i=1:n$, as
\begin{equation}
\frac{\partial J}{\partial x_i^t}=\sum_{j=1}^{n}\frac{\partial J}{\partial z_j^{t+1}} \frac{\partial z_j^{t+1}}{\partial x_i^t}+\sum_{j=1}^{n}\frac{\partial J}{\partial z_j^{t+2}} \frac{\partial z_j^{t+2}}{\partial x_i^t},
\label{derivation_step1}
\end{equation}
since $\bold x^t$ influences the cost $J$ via $\bold z^{t+1}$ and $\bold z^{t+2}$. Now, using the relations $\mathbf{z}^{t+1} = (1+\beta_{t+1}) \mathbf{x}^{t} - \beta_{t+1} \mathbf{x}^{t-1}$ and $\mathbf{z}^{t+2} = (1+\beta_{t+2}) \mathbf{x}^{t+1} - \beta_{t+2} \mathbf{x}^{t}$, we observe that $\frac{\partial z_j^{t+1}}{\partial x_i^t}=\frac{\partial z_j^{t+2}}{\partial x_i^t}=0$, whenever $i \neq j$; and
\begin{equation*}
\frac{\partial z_i^{t+1}}{\partial x_i^t}=1+\beta_{t+1} \text{\,\,and\,\,} \frac{\partial z_i^{t+2}}{\partial x_i^t}=-\beta_{t+2}, \text{\,\,for all\,\,}i.
\end{equation*}
Therefore, a vectorized representation of \eqref{derivation_step1} leads to
\begin{align}
\label{eq: dJ_dx_value}
	\nabla_{\mathbf{x}^t}J = (1+\beta_{t+1})\, \nabla_{\mathbf{z}^{t+1}}J - \beta_{t+2} \, \nabla_{\mathbf{z}^{t+2}}J, \text{\,\,where\,\,} t=1:L-2.  
\end{align}
Corresponding to $t=L-1$ and $t=L$, we have $\nabla_{\mathbf{x}^{L-1}}J = (1+\beta_{L})\, \nabla_{\mathbf{z}^{L}}J$ and $\nabla_{\mathbf{x}^L}J = \mathbf{x}^L - \mathbf{x}$, respectively.\\
\indent To evaluate $\frac{\partial J}{\partial z_j^{t}}$, we use the chain-rule of differentiation:
\begin{equation}
\frac{\partial J}{\partial z_j^{t}}=\sum_{\ell=1}^{n}\frac{\partial J}{\partial \tilde{x}_{\ell}^{t}}  \frac{\partial \tilde{x}_{\ell}^{t}}{\partial z_j^{t}}=\sum_{\ell=1}^{n}\frac{\partial \tilde{x}_{\ell}^{t}}{\partial z_j^{t}}\sum_{i=1}^{n}\frac{\partial J}{\partial x_i^{t}}  \frac{\partial x_i^{t}}{\partial \tilde{x}_{\ell}^{t}}.
\label{derivation_step2}
\end{equation}
Further, using the facts that $\tilde{\mathbf{x}}^{t} = \mathbf{W}\mathbf{z}^t + \mathbf{b}$ and $\bold x^t=\psi^{(t)}\left(\tilde{\bold x}^t  \right)$, where the activation $\psi^{(t)}$ in the $t^{\text{th}}$ layer is applied element-wise, we represent \eqref{derivation_step2} compactly as
\begin{align}
\label{eq: dJ_dzt_value}
	\nabla_{\mathbf{z}^{t}}J = \mathbf{W}^\top\text{diag}\left(\psi^{'(t)}\left(\tilde{\mathbf{x}}^t\right)\right) \nabla_{\mathbf{x}^t}J.
\end{align}
The steps for computing the gradient for the \textit{fLETnet} architecture using the back-propagation algorithm are summarized below:
\begin{enumerate}
\item $\nabla_{\mathbf{z}^{t}}J = \mathbf{W}^\top\text{diag}\left(\psi^{'(t)}\left(\tilde{\mathbf{x}}^t\right)\right) \nabla_{\mathbf{x}^t}J$, for $t=L$ down to $1$;

\item $\nabla_{\mathbf{x}^t}J = (1+\beta_{t+1})\, \nabla_{\mathbf{z}^{t+1}}J - \beta_{t+2} \, \nabla_{\mathbf{z}^{t+2}}J$, for $t=L-2$ down to $1$; $\nabla_{\mathbf{x}^{L-1}}J = (1+\beta_{L})\, \nabla_{\mathbf{z}^{L}}J$; and
\item $\nabla_{\mathbf{c}^t}J = (\mathbf{\Phi}^t)^\top \nabla_{\mathbf{x}^t}J$, for $t=L$ down to $1$.
\end{enumerate}
We use $\nabla_{\mathbf{x}^L}J = \mathbf{x}^L - \mathbf{x}$ to initialize the algorithm. The variables $\bold x^t$ and $\tilde{\bold x}^t=\bold W \bold z^t+\bold b$, for $t=1:L$, are calculated and stored in a forward pass through the \textit{fLETnet}.
\section{Hessian-Vector Product Computation for \textit{fLETnet}}
\label{supp3}
Analogous to \textit{LETnetVar}, the Hessian-vector product $\bold H \bold u$, where $\bold H=\nabla^2 J(\bold c)$ and $\bold u$ is any vector having the same dimension as $\bold c$, can be exactly computed as 
\begin{align}
	\mathbf{Hu} = \mathcal{R}_{\mathbf{u}}\left(\nabla_{\mathbf{c}} J\right),
\end{align}
where the operator $\mathcal{R}_{\mathbf{u}}$ has been defined in (M-27). To evaluate $\mathcal{R}_{\mathbf{u}}\left(\nabla_{\mathbf{c}} J\right)$, one must run a forward pass and a backward pass through the \textit{fLETnet}.
\subsection{Forward pass}
Applying $\mathcal{R}_{\bold u}$ on both sides of $\mathbf{z}^{t} = (1+\beta_{t}) \mathbf{x}^{t-1} - \beta_{t} \mathbf{x}^{t-2}$ and $\tilde{\bold x}^t=\bold W \bold z^t+\bold b$, we obtain
\begin{align}
	\mathcal{R}_{\bold u}(\mathbf{z}^{t}) &= (1+\beta_t) \mathcal{R}_{\bold u}(\mathbf{x}^{t-1}) - \beta_t \mathcal{R}_{\bold u}(\mathbf{x}^{t-2}), \text{\,and} \label{eq: R_forward_pass_fLETnet_1}\\
	\mathcal{R}_{\bold u}(\tilde{\mathbf{x}}^{t}) &= \mathbf{W}\mathcal{R}_{\bold u}(\mathbf{z}^t)=(1+\beta_{t})\bold W \mathcal{R}_{\bold u}\left(\mathbf{x}^{t-1} \right)- \beta_{t} \bold W \mathcal{R}_{\bold u}\left(\mathbf{x}^{t-2}\right), \label{eq: R_forward_pass_fLETnet_2}
\end{align}
respectively; using Properties 1, 3, and 4 of $\mathcal{R}_{\bold u}$ listed in Appendix B. Evaluation of \eqref{eq: R_forward_pass_fLETnet_2} requires computation of $\mathcal{R}_{\bold u}\left(\mathbf{x}^{t}\right)$, for $t=1:L$. Similar to (M-30), we apply $\mathcal{R}_{\bold u}$ on both sides of $\bold x^t=\psi^{(t)}\left(\tilde{\bold x}^t\right)=\sum_{k=1}^{K} c^t_k \phi_k(\tilde{\mathbf{x}}^t)$ to obtain
\begin{align}
	\mathcal{R}_{\bold u}(\mathbf{x}^t) &= \sum_{k=1}^{K} \left[ u^t_k \phi_k(\tilde{\mathbf{x}}^t) + c^t_k \mathcal{R}_{\bold u}(\phi_k(\tilde{\mathbf{x}}^t))\right],
\end{align}
where $\mathcal{R}_{\bold u}(\phi_k(\tilde{\mathbf{x}}^t))=\text{diag}\left(\phi_k^{'}\left(\tilde{\bold x}^t\right)\right) \mathcal{R}_{\bold u}\left(\tilde{\mathbf{x}}^t \right)$, as given in (M-32). The quantities $\mathcal{R}_{\bold u}\left(\tilde{\mathbf{x}}^t \right)$ and $\mathcal{R}_{\bold u}\left(\mathbf{x}^t \right)$ are computed iteratively and stored in the forward pass, for $t=1:L$:
\begin{enumerate}
\item $\mathcal{R}_{\bold u}\left(\tilde{\mathbf{x}}^t \right)=(1+\beta_{t})\bold W \mathcal{R}_{\bold u}\left(\mathbf{x}^{t-1} \right)- \beta_{t} \bold W \mathcal{R}_{\bold u}\left(\mathbf{x}^{t-2}\right)$;
\item $\mathcal{R}_{\bold u}(\phi_k(\tilde{\mathbf{x}}^t))=\text{diag}\left(\phi_k^{'}\left(\tilde{\bold x}^t\right)\right) \mathcal{R}_{\bold u}\left(\tilde{\mathbf{x}}^t \right)$, for $k = 1:K$; and
\item $\mathcal{R}_{\bold u}\left({\mathbf{x}}^t \right)=\sum_{k=1}^{K} \left[ u^t_k \phi_k(\tilde{\mathbf{x}}^t) + c^t_k\mathcal{R}_{\bold u}(\phi_k(\tilde{\mathbf{x}}^t))\right]$;
\end{enumerate}
where $\mathcal{R}_{\bold u}\left({\mathbf{x}}^0 \right)=\mathcal{R}_{\bold u}\left({\mathbf{x}}^{-1} \right)=\bold 0$, as the initializations ${\mathbf{x}}^0$ and ${\mathbf{x}}^{-1}$ are independent of $\bold c$.
\subsection{Backward pass}
Applying $\mathcal{R}_{\bold u}$ on \eqref{eq: dJ_dct_value}, we get
\begin{align}
\label{eq: R_dJ_dct}
\hspace{-2mm}	\mathcal{R}_{\bold u}(\nabla_{\mathbf{c}^t}J) = \left( \boldsymbol \Phi^t\right)^\top \mathcal{R}_{\bold u}\left( \nabla_{\bold x^t}J \right)+\left(\mathcal{R}_{\bold u} \left( \boldsymbol \Phi^t\right)\right)^\top \nabla_{\bold x^t}J,
\end{align}
employing an argument similar to the one used to obtain (M-35). Further, applying $\mathcal{R}_{\bold u}$ on (\ref{eq: dJ_dx_value}) leads to
\begin{align}
\label{eq: R_dJ_dx}
	\hspace{-3mm}\mathcal{R}_{\bold u}(\nabla_{\mathbf{x}^t}J) = (1+\beta_{t+1})\mathcal{R}_{\bold u}(\nabla_{\mathbf{z}^{t+1}}J) - \beta_{t+2} \mathcal{R}_{\bold u}( \nabla_{\mathbf{z}^{t+2}}J),
\end{align}
where $t=L-2$ down to $1$. For layers $t=L-1$ and $t=L$, we have 
\begin{eqnarray}
\mathcal{R}_{\bold u}\left(\nabla_{\mathbf{x}^{L-1}}J \right) &=& (1+\beta_{L})\, \mathcal{R}_{\bold u}(\nabla_{\mathbf{z}^{L}}J), \text{\,\,and}\\
\mathcal{R}_{\bold u}(\nabla_{\mathbf{x}^L}J) &=& \mathcal{R}_{\bold u}(\mathbf{x}^L),
\end{eqnarray}
respectively, where $\mathcal{R}_{\bold u}(\mathbf{x}^L)$ is computed in the forward pass through \textit{fLETnet}. To evaluate $ \mathcal{R}_{\bold u}( \nabla_{\mathbf{z}^{t}}J)$, we apply $ \mathcal{R}_{\bold u}$ on both sides of
\begin{equation*}
\nabla_{\mathbf{z}^{t}}J = \mathbf{W}^\top\text{diag}\left(\psi^{'(t)}\left(\tilde{\mathbf{x}}^t\right)\right) \nabla_{\mathbf{x}^t}J,
\end{equation*}
as obtained from Step 1 of the forward pass, resulting in
\begin{eqnarray}
\label{eq: R_dJ_dz}
\mathcal{R}_{\bold u}(\nabla_{\mathbf{z}^{t}}J) &=& \mathbf{W}^\top \Bigg[
\text{diag}\left(\mathcal{R}_{\bold u}\left(\psi^{'(t)}\left(\tilde{\mathbf{x}}^t\right)\right)\right)  \nabla_{\mathbf{x}^t}J\nonumber\\
&&+ \text{diag}\left(\psi^{'(t)}\left(\tilde{\mathbf{x}}^t\right)\right)  \mathcal{R}_{\bold u}(\nabla_{\mathbf{x}^t}J )\Bigg],
\end{eqnarray}
using an argument similar to that used to arrive at (M-41). The quantity $\mathcal{R}_{\bold u}\left(\psi^{'(t)}\left(\tilde{\bold x}^{t}\right)  \right)$ can be computed using (M-39) and (M-40). The back-propagation algorithm to compute the Hessian-vector product is summarized below:
\begin{enumerate}
\item $\mathcal{R}_{\bold u}(\nabla_{\mathbf{z}^{t}}J) = \mathbf{W}^\top \Big[
\text{diag}\left(\mathcal{R}_{\bold u}\left(\psi^{'(t)}\left(\tilde{\mathbf{x}}^t\right)\right)\right)  \nabla_{\mathbf{x}^t}J  + \text{diag}\left(\psi{'(t)}\left(\tilde{\mathbf{x}}^t\right)\right)  \mathcal{R}_{\bold u}(\nabla_{\mathbf{x}^t}J )\Big]$, for $t=L$ down to $1$;

\item $\mathcal{R}_{\bold u}(\nabla_{\mathbf{x}^t}J) = (1+\beta_{t+1})\mathcal{R}_{\bold u}(\nabla_{\mathbf{z}^{t+1}}J) - \beta_{t+2} \mathcal{R}_{\bold u}( \nabla_{\mathbf{z}^{t+2}}J)$, for $t=L-2$ down to $1$; $\mathcal{R}_{\bold u}\left(\nabla_{\mathbf{x}^{L-1}}J \right) = (1+\beta_{L})\, \mathcal{R}_{\bold u}(\nabla_{\mathbf{z}^{L}}J)$; and $\mathcal{R}_{\bold u}(\nabla_{\mathbf{x}^L}J) = \mathcal{R}_{\bold u}(\mathbf{x}^L)$;

\item $\mathcal{R}_{\bold u}(\nabla_{\mathbf{c}^t}J) = \left( \boldsymbol \Phi^t\right)^\top \mathcal{R}_{\bold u}\left( \nabla_{\bold x^t}J \right)+\left(\mathcal{R}_{\bold u} \left( \boldsymbol \Phi^t\right)\right)^\top \nabla_{\bold x^t}J$, for $t=L$ down to $1$.
\end{enumerate}
\section{Regularizers Learnt by a \textit{fLETnet}}
\label{supp4}
\indent In Section 7.3 and Figure 6, we showed the activations learnt, the corresponding regularizers, and a portion of the estimated signal, only for layers 8 -- 14 for want of space. The evolution of the activation functions, corresponding regularizers, and the estimated signals across $L=50$ layers of a trained \textit{fLETnet} are shown in the following figure. The layer indices are indicated at the top of every block. We observe that the learnt activations and the corresponding regularizers evolve in such a way that a balance is maintained between noise cancellation and signal preservation. The estimated signal entry in one layer might increase in the subsequent one, as elaborated in Section 7.3 of the manuscript, due to flexible parametric design of the regularizers using the LET.

\begin{figure*}[htpb]		
	\captionsetup[subfigure]{labelformat=empty}
\begin{subfigure}{\textwidth}
		\centering
		\includegraphics[]{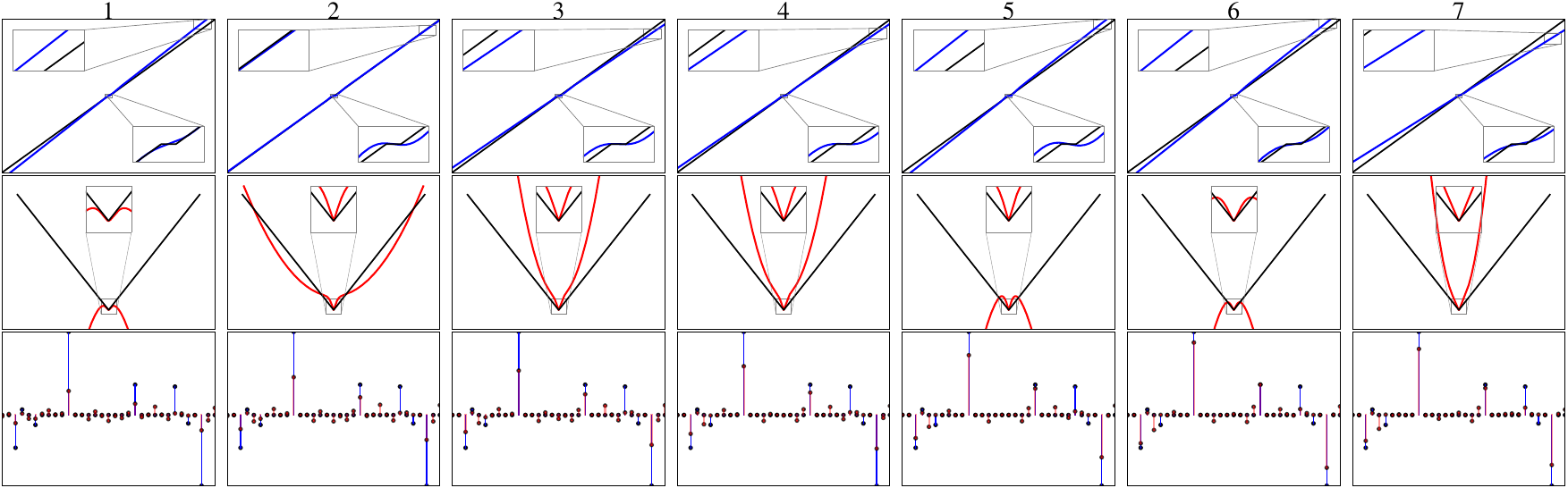}
	\end{subfigure}\\[2mm]
	\begin{subfigure}{\textwidth}
		\centering
		\includegraphics[]{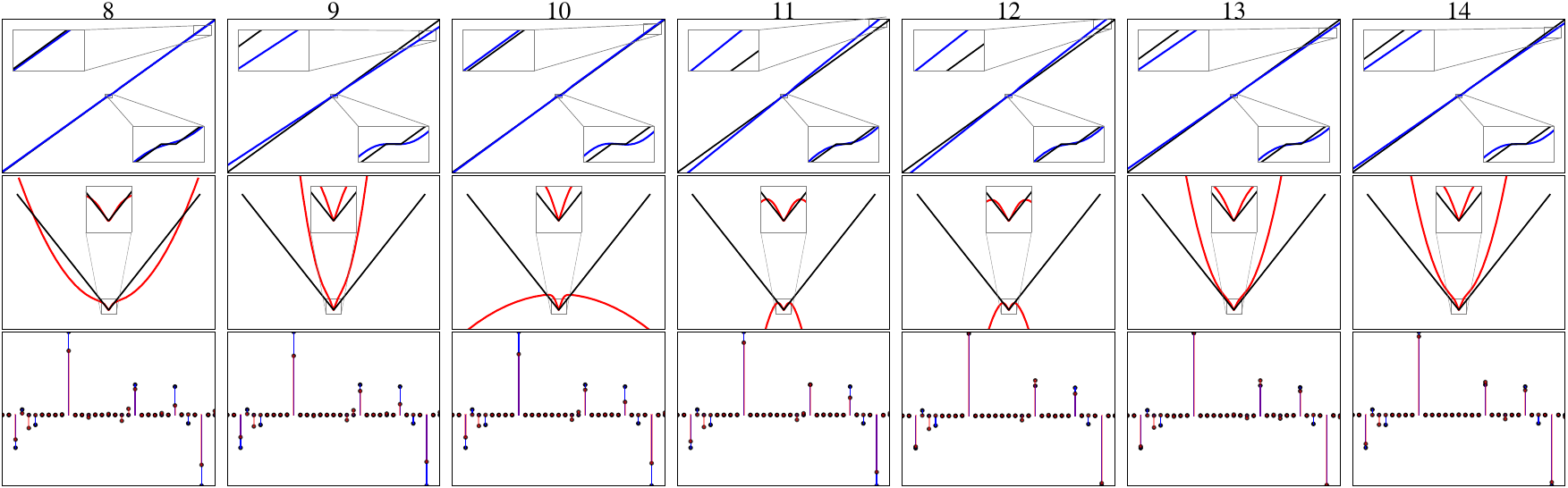}
	\end{subfigure} \\[2mm]
	\begin{subfigure}{\textwidth}
		\centering
		\includegraphics[]{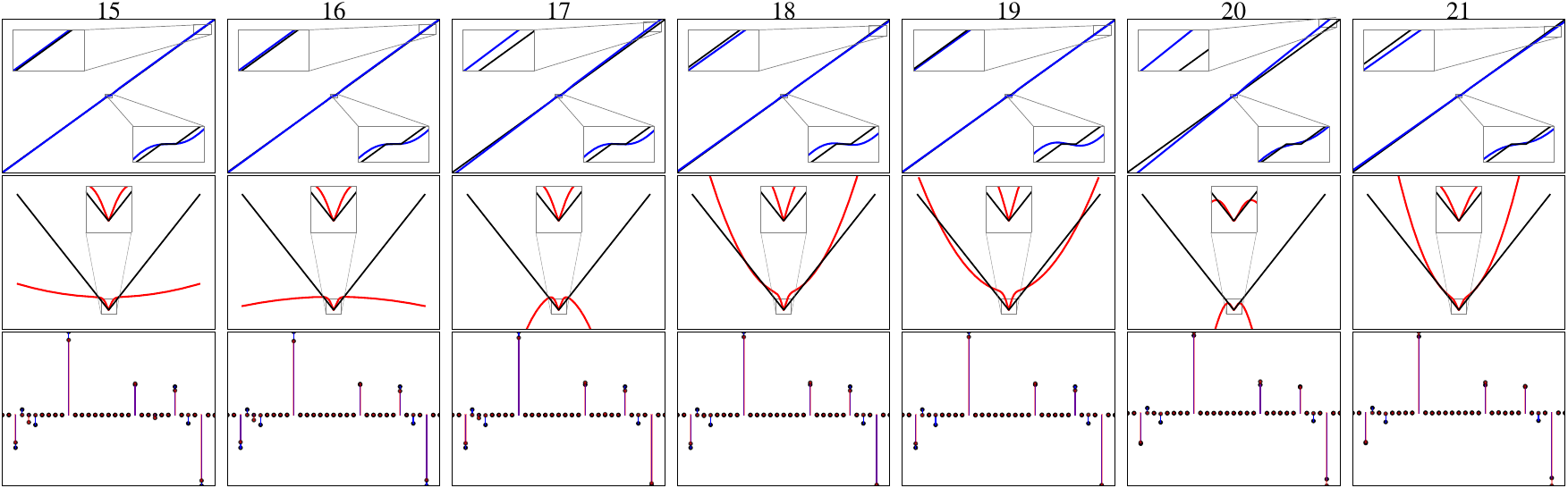}
	\end{subfigure} \\[2mm]
	\begin{subfigure}{\textwidth}
		\centering
		\includegraphics[]{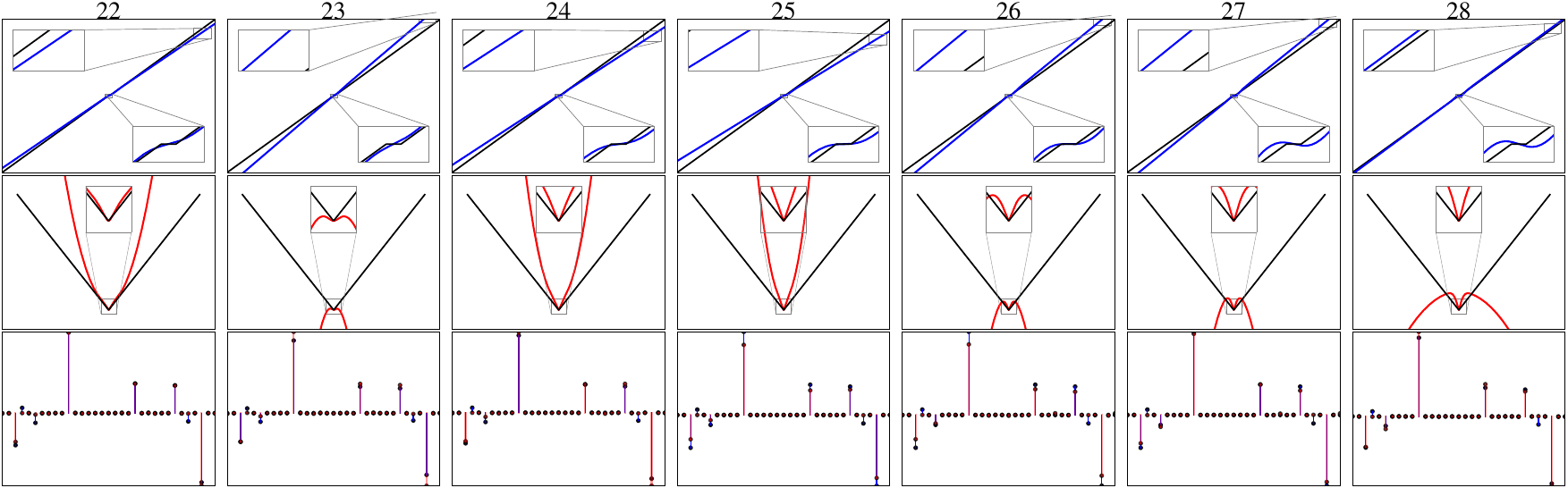}
	\end{subfigure}\\[2mm]
	\caption{}
\end{figure*}

\begin{figure*}[htpb]		
	\captionsetup[subfigure]{labelformat=empty}
		
	\begin{subfigure}{\textwidth}
		\centering
		\includegraphics[]{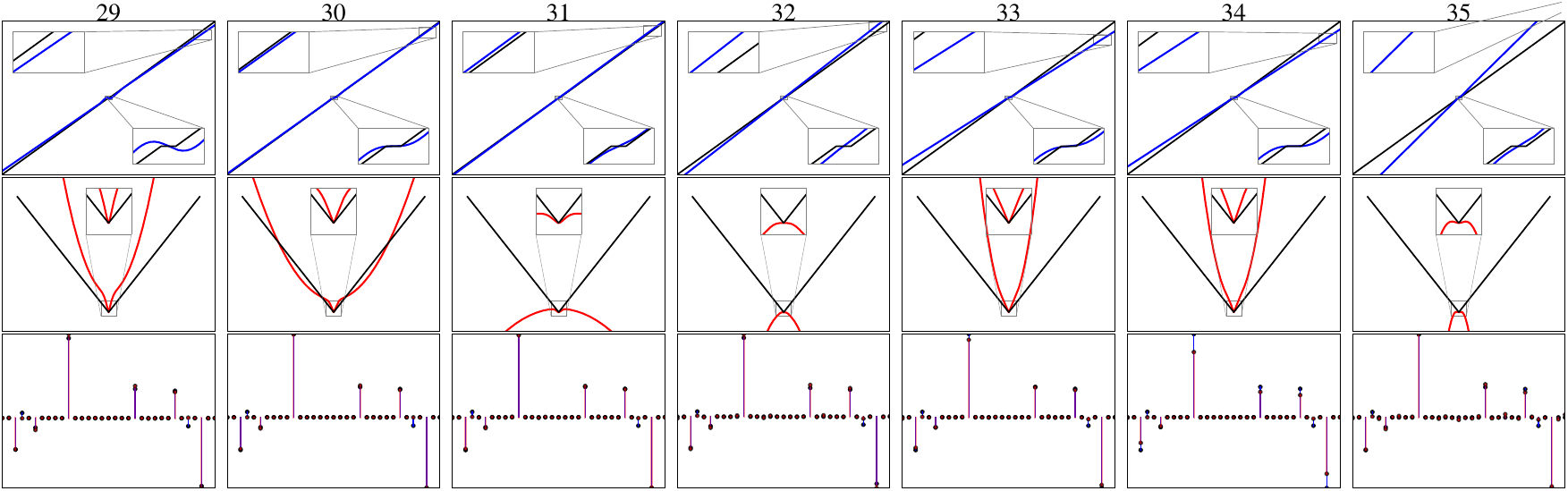}
	\end{subfigure}\\[2mm]
	\begin{subfigure}{\textwidth}
		\centering
		\includegraphics[]{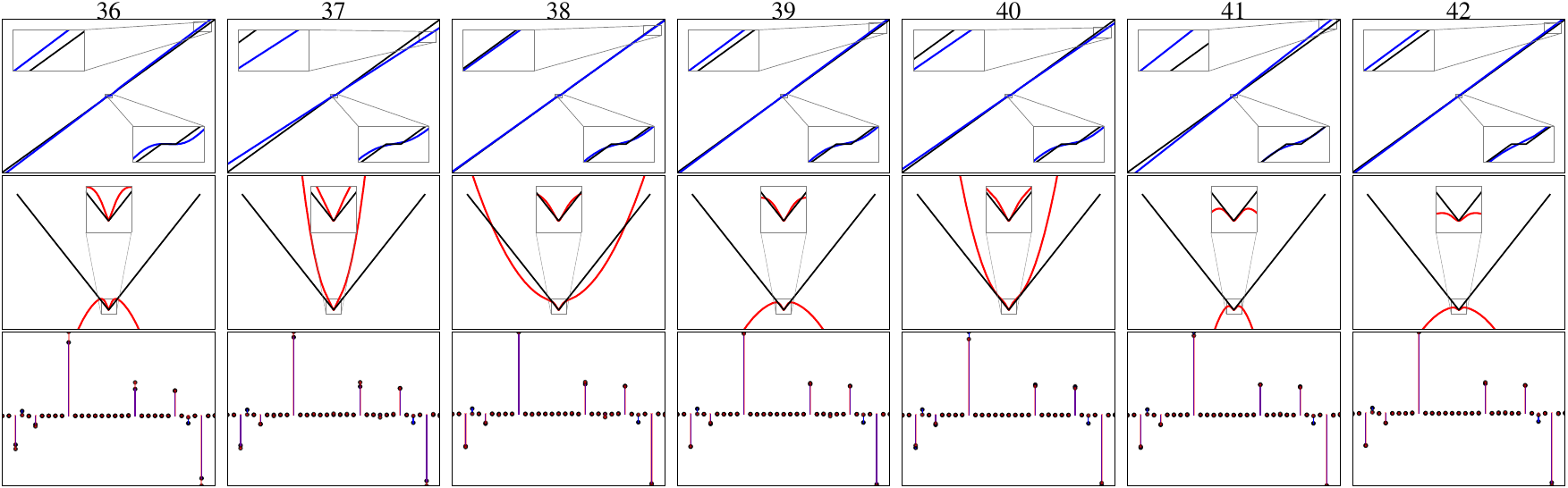}
		\begin{subfigure}{\textwidth}
		\centering
		\includegraphics[]{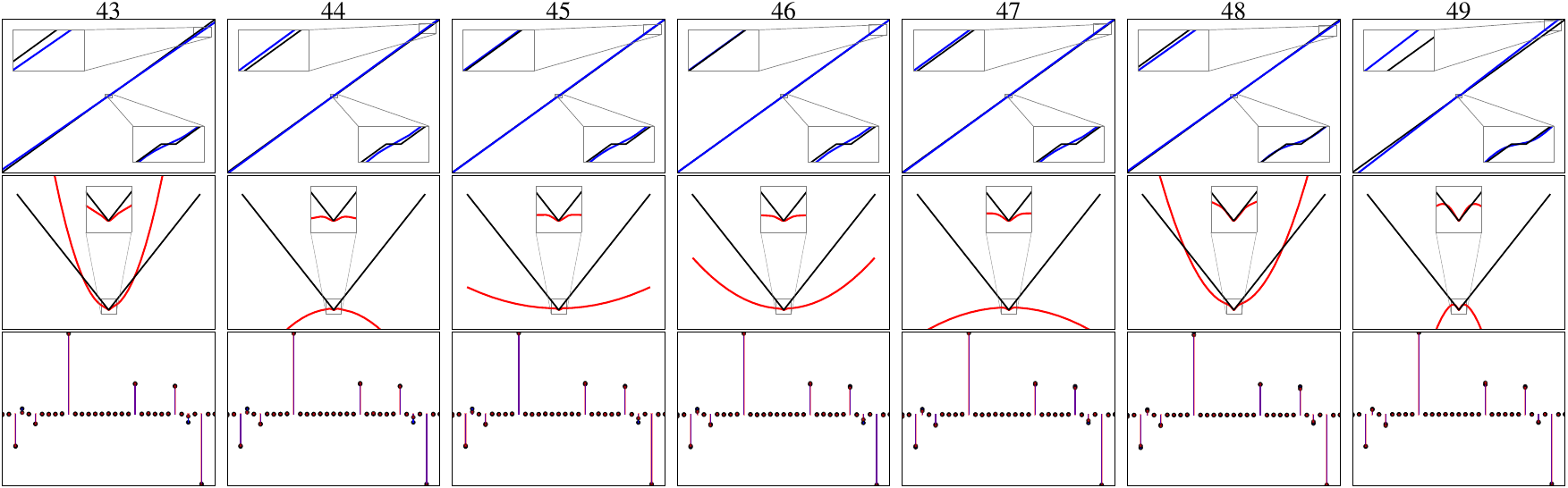}
	\end{subfigure} \\[2mm]
	\begin{subfigure}{\textwidth}
		\centering
		\includegraphics[]{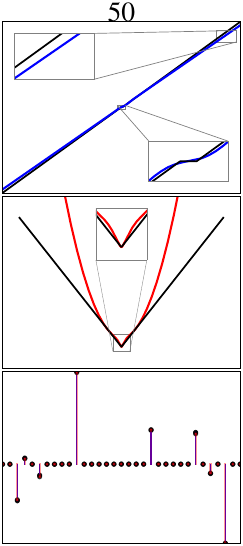}
	\end{subfigure}
	\end{subfigure}
	\caption{The learnt parametric LET functions (blue) of \textit{fLETnet} over a typical dataset and their corresponding induced regularizers (red). These functions compared against ST and $\ell_1$ regularizer (black). The number on top of each figure is the layer index of the network.}
	\label{fig: Learnt_regularizers}
\end{figure*}
\end{document}